\documentclass[11pt]{article}

\usepackage[final]{acl}

\usepackage{times}
\usepackage{latexsym}

\usepackage[T1]{fontenc}

\usepackage[utf8]{inputenc}

\usepackage{microtype}

\usepackage{inconsolata}

\usepackage{hyperref}
\usepackage{url}
\usepackage{soul}
\usepackage{subcaption}
\usepackage{amsmath}
\usepackage{wasysym}
\usepackage{makecell}
\usepackage{graphicx}
\usepackage[listings,breakable,skins]{tcolorbox}
\usepackage{booktabs}
\usepackage{multirow}
\usepackage{threeparttable}
\usepackage{adjustbox}
\usepackage[normalem]{ulem}
\usepackage{wrapfig}
\newtcblisting{prompttemplate}[1]{
  title={#1},
  breakable,
  boxrule=0.5mm,
  listing only,
  listing options={
    basicstyle=\ttfamily,
    breaklines=true,
    columns=fullflexible,
    keepspaces=true,
  }
}
\newtcblisting{prompttemplateWhite}[1]{
  title={#1},
  breakable,
  boxrule=0.5mm,
  colframe=gray!50!black,
  colback=white,
  coltitle=white, 
  fonttitle=\scshape,
  listing only,
  listing options={
    basicstyle=\ttfamily\footnotesize,
    breaklines=true,
    columns=fullflexible,
    keepspaces=true,
  }
}

\newtcblisting{prompttemplateBlue}[1]{
  title={#1},
  breakable,
  boxrule=0.5mm,
  colframe=blue!50!black,
  colback=white,
  coltitle=white, 
  fonttitle=\scshape,
  listing only,
  listing options={
    basicstyle=\ttfamily\footnotesize,
    breaklines=true,
    columns=fullflexible,
    keepspaces=true,
  }
}

\newtcblisting{prompttemplateBlue*}[1]{
  title={#1},
  boxrule=0.5mm,
  colframe=blue!50!black,
  colback=white,
  coltitle=white,
  fonttitle=\scshape,
  listing only,
  listing options={
    basicstyle=\ttfamily\footnotesize,
    breaklines=true,
    columns=fullflexible,
    keepspaces=true,
  },
  float*=!t,
  width=\textwidth
}

\newtcblisting{prompttemplateOrange}[1]{
  title={#1},
  breakable,
  boxrule=0.5mm,
  colframe=orange!50!black,
  colback=white,
  coltitle=white, 
  fonttitle=\scshape,
  listing only,
  listing options={
    basicstyle=\ttfamily\footnotesize,
    breaklines=true,
    columns=fullflexible,
    keepspaces=true,
  }
}

\title{\textsc{OptSkills}: Learning Generalizable Optimization Skills from Problem Archetypes via Cluster-Based Distillation}

\author{
    Haochen Yang$^{1}$\thanks{These authors contributed equally to this work.}, Ke Zhao$^{1\ast}$, Mengyuan Ma$^{1}$, Xingyu Lu$^{2}$, Xiangfeng Wang$^{4,5}$, Hong Qian$^{1,3}$ \thanks{Corresponding author}\\
    $^{1}$Shanghai Institute of AI for Education and School of Computer Science and Technology,\\ East China Normal University, Shanghai, China \\
    $^{2}$Ant Group, Hangzhou, China \\
    $^{3}$Shanghai Innovation Institute, Shanghai, China \\
    $^{4}$Key Laboratory of Mathematics and Engineering Applications (MoE) and School of \\Mathematical Sciences, East China Normal University, Shanghai, China \\
    $^{5}$Shenzhen Loop Area Institute, Shenzhen, China \\
    \texttt{\{51290936033,kezhao,10222140418\}@stu.ecnu.edu.cn}, \\
    \texttt{sing.lxy@antgroup.com,xfwang@sei.ecnu.edu.cn,hqian@cs.ecnu.edu.cn}
}

\begin{document}
\maketitle
\begin{abstract}
Leveraging Large Language Models (LLMs) to automatically formulate and solve optimization problems from natural language has emerged as an efficient paradigm for automated optimization. However, existing methods still exhibit limited generalization: they are sensitive to superficial narrative variations, reuse experience mainly at the case level, and struggle to adapt to shifted or emerging problem types. We propose \textsc{OptSkills}, an archetype-centric skill learning and reasoning agent system for optimization modeling and solving. To improve robust generalization, our system clusters problems by their underlying archetypes rather than surface narratives. To improve in-distribution generalization, it explores diverse modeling paradigms and solver configurations within each cluster, then distills successful trajectories into reusable workflow-level skills. To improve out-of-distribution generalization, it refines existing skills or expands the skill library using newly obtained trajectories. Our system achieves a state-of-the-art micro-averaged accuracy of 68.27\% on datasets encompassing diverse problem types and scenarios. In addition, on MIPLIB-NL, a highly challenging large-scale and high-dimensional benchmark, it achieves 26.91\% accuracy, outperforming DeepSeek-V3.2-Thinking by 4.53\%. After skill learning on NANO-CO, it reaches 72.79\% on the OOD NLCO benchmark.
Code and skills are available at \href{https://github.com/fujiwaranoM0kou/OptSkills}{https://github.com/fujiwaranoM0kou/OptSkills}.

\end{abstract}

\begin{figure}[!h]
    \centering

    \begin{subfigure}{0.48\columnwidth}
        \centering
        \includegraphics[width=\columnwidth]{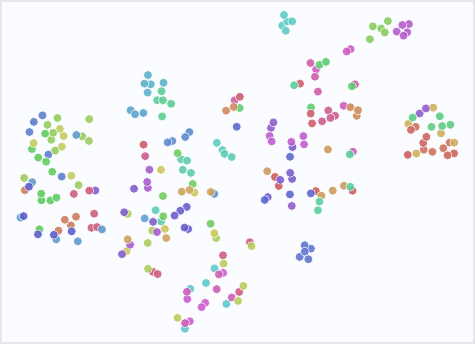}
        \caption{Raw problem embedding.}
        \label{fig:a}
    \end{subfigure}
    \hfill
    \begin{subfigure}{0.48\columnwidth}
        \centering
        \includegraphics[width=\columnwidth]{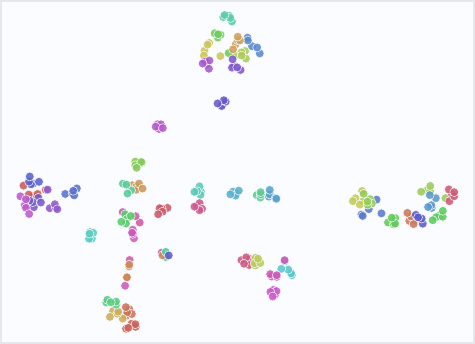}
        \caption{Archetype embedding.}
        \label{fig:b}
    \end{subfigure}

    \begin{subfigure}{0.48\columnwidth}
        \centering
        \includegraphics[width=\columnwidth]{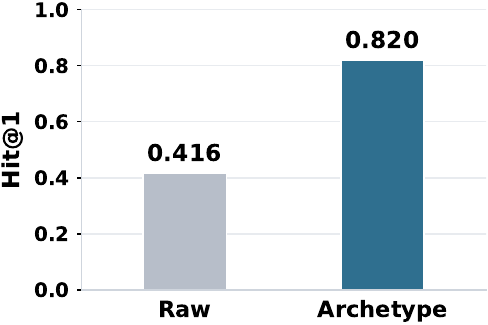}
        \caption{Hit@1.}
        \label{fig:c}
    \end{subfigure}
    \hfill
    \begin{subfigure}{0.48\columnwidth}
        \centering
        \includegraphics[width=\columnwidth]{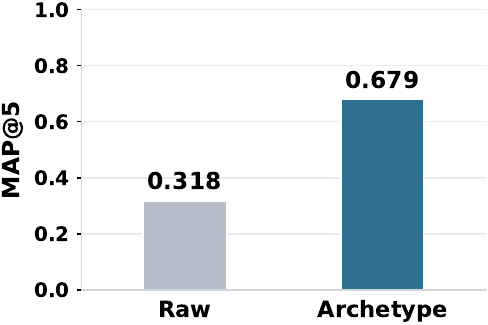}
        \caption{MAP@5.}
        \label{fig:d}
    \end{subfigure}

    \caption{Embedding-space comparison on \textsc{NANO-CO}.
    Each point represents a problem instance, colored by manually verified problem-archetype labels. 
    Top panels show $t$-SNE projections of raw problem-text embeddings and archetype embeddings; bottom panels report nearest-neighbor retrieval results in Hit@1 and MAP@5. Details are provided in Appendix~\ref{appendix_knn}.}
    \label{fig:four}
\end{figure}
\section{Introduction}
The advancement of the industrial field brings the efficient solution of operational optimization problems into sharp focus across academia and industry. However, the inherent complexity of optimization problems has necessitated reliance on domain experts to formulate mathematical models and develop solver code, a process that is both costly and expertise-intensive. Large language models (LLMs) have recently been explored as a promising tool for automating optimization modeling and solving. Existing studies have investigated multi-agent collaboration~\cite{pmlr-v235-ahmaditeshnizi24a, xiao2024chainofexperts}, supervised fine-tuning (SFT)~\cite{JiangShu2025llmopt, Huang2025}, and reinforcement learning~\cite{chen2026solverinformed,zhao2026minioptreasoningmodelsolve}, achieving strong performance on standard benchmarks.

Recent studies have explored experience reuse for LLM-based optimization. AlphaOPT~\cite{kong2025alphaopt} constructs a self-evolving experience library that retrieves relevant experiences to guide similar problems, while LEAN-LLM-OPT~\cite{liang2026largescaleoptimizationmodelautoformulation} introduces a library of few-shot examples to help LLM agents construct structured workflows. However, these agents still suffer from limited generalization ability. In practical scenarios, this limitation appears in three forms. First, \textbf{\textit{robust generalization}} requires the agent to identify the same optimization structure under different narratives, while avoiding sensitivity to superficial textual variations. Second, \textbf{\textit{in-distribution (ID) generalization}} requires the agent to solve new instances of known optimization archetypes by reusing stable modeling and solving workflows. Third, \textbf{\textit{out-of-distribution (OOD) generalization}} requires the agent to adapt when new domains introduce semantic variants of known archetypes or entirely new archetypes. These methods mainly organize reusable knowledge at the case level. Such case-level reuse can be sensitive to narrative changes, insufficient for distilling workflow-level knowledge shared by structurally equivalent problems, and limited in adapting to shifted or emerging problem types. This motivates an \textbf{\textit{archetype-centric view}} of experience reuse, where optimization problems are organized by their canonical structures rather than surface textual similarity.

To provide empirical analysis for this view, this paper conducts exploratory experiments on the embedding-space clusterability of optimization problems under archetype-based representations. We use problems from the combinatorial optimization dataset \textsc{NANO-CO}, a self-constructed dataset described in Appendix~\ref{appendix_datasynthesis}. As shown in Figure~\ref{fig:b}, when problems are represented by their archetype embeddings, instances with similar optimization structures form compact and well-separated groups under $t$-SNE~\cite{van2008visualizing}. In contrast, Figure~\ref{fig:a} shows that raw problem-text embeddings lead to lower intra-class compactness and severe inter-class overlap, indicating that surface narratives are not reliable indicators of reusable workflows. Figures~\ref{fig:c} and~\ref{fig:d} further show that archetype embeddings improve nearest-neighbor retrieval in both Hit@1 and MAP@5. These results suggest that archetype-based representations provide a more suitable foundation for retrieving, clustering, and acquiring reusable optimization experience.

Based on the observations above, this paper proposes \textsc{OptSkills}, an agent system for archetype-centric workflow reuse in optimization modeling and solving. \textit{To address the brittleness of case-level reuse under narrative variations}, \textsc{OptSkills} clusters optimization problems according to their underlying archetypes rather than surface-level textual similarity, allowing semantically different descriptions with the same canonical structure to share reusable experience. \textit{To improve ID generalization within known optimization archetypes}, \textsc{OptSkills} explores diverse modeling paradigms and solver configurations inside each archetype cluster, and distills intra-cluster trajectories into reusable workflow-level skills.  \textit{Archetype-based cluster-level skills already exhibit OOD generalization capabilities. To continuously enhance the optimization generalization of \textsc{OptSkills} on OOD samples}, we introduce skill learning, which supports learning skills from newly obtained feasible trajectories, either refining existing skills for semantic variants of known archetypes or expanding the skill library for emerging problem types. In this way, \textsc{OptSkills} contributes an archetype-centric experience reuse framework that directly targets robust generalization across narratives, ID generalization within known archetypes, and OOD adaptation to shifted optimization scenarios. 

We evaluate \textsc{OptSkills} on problem datasets encompassing multiple problem types and scenarios, where it achieves a state-of-the-art Micro-Avg. accuracy of \textit{68.27\%}, outperforming the strongest baseline Trace2Skill by \textit{4.81\%}. On the more challenging large-scale MIPLIB-NL benchmark, \textsc{OptSkills} achieves \textit{26.91\%} accuracy, outperforming DeepSeek-V3.2-Thinking by \textit{4.53\%}. After skill learning on the self-constructed \textsc{NANO-CO} dataset, \textsc{OptSkills} further reaches \textit{72.79\%} accuracy on the OOD NLCO benchmark, showing the benefit of archetype-centric experience reuse under distribution shifts.


The remainder of this paper is organized as follows. Section~2 reviews related work. Section~3 presents the proposed \textsc{OptSkills} framework. Section~4 describes the experimental setup and analyzes the results. Section~5 concludes the paper.

\begin{figure*}[t]
    \centering
    \includegraphics[width=\textwidth]{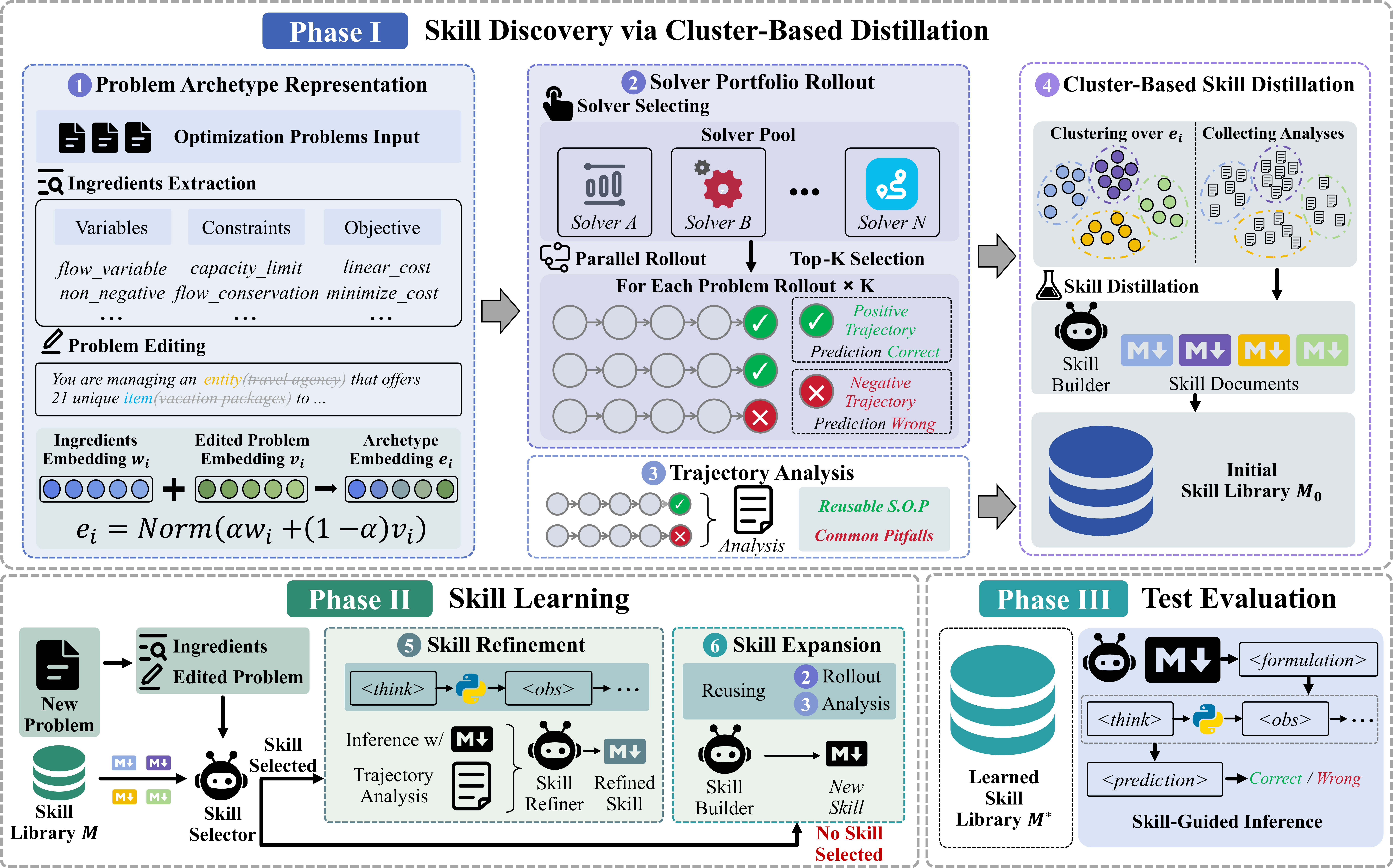}
\caption{
\textbf{An overview of \textsc{OptSkills}.}
\textsc{OptSkills} consists of three phases.
\textbf{Phase I:} \textbf{(1) Problem Archetype Representation}, \textbf{(2) Solver Portfolio Rollout}, \textbf{(3) Trajectory Analysis}, and \textbf{(4) Cluster-Based Skill Distillation} build the initial skill library $\mathcal{M}_0$.
\textbf{Phase II:} for a new problem, skill selector routes it to \textbf{(5) Skill Refinement} or \textbf{(6) Skill Expansion}, yielding the learned skill library $\mathcal{M}^{\ast}$.
\textbf{Phase III:} the selected skill guides test-time inference.
}
\label{fig:framework}
\end{figure*}

\section{Related Work}
\label{relatedwork}
\subsection{LLMs for Modeling and Solving Optimization Problems}

\paragraph{Post-Training Methods.} These approaches seek to encode optimization modeling directly into mathematical models through post-training of LLMs. Representative studies are based on multiple paradigms such as Supervised Fine-tuning (ORLM~\cite{Huang2025}, OptMATH~\cite{lu2025optmathscalablebidirectionaldata}, AutoOR~\cite{motwani2026autoor}), Preference Optimization (LLMOPT~\cite{JiangShu2025llmopt}), Reinforcement Learning (SIRL~\cite{chen2026solverinformed}, MiniOpt\cite{zhao2026minioptreasoningmodelsolve},OR-R1~\cite{ding2026or}), and SFT combined with evolutionary computation(LLaMoCo~\cite{11359290}).

\textbf{Agent-Based Methods.} Rather than updating model parameters, these methods orchestrate multi-stage LLM workflows that decompose modeling, coding, solving, and verification into modular stages. Chain-of-Experts~\cite{xiao2024chainofexperts} and OptiMUS~\cite{pmlr-v235-ahmaditeshnizi24a} partition complex problems across specialized expert agents and organize them into structured agent topologies. ORMind~\cite{wang2025ormind} and ORThought~\cite{yang2025orthought} incorporate cognitive reasoning and expert-level modeling heuristics to systematically map business requirements onto mathematical abstractions and solver implementations. 

\textbf{Experience-Enhanced Methods.} Recent efforts begin integrating external experience, shifting the LLM-based optimization tasks from {single-shot generation} toward {experience-augmented modeling}. DRoC~\cite{jiang2025droc} and ConstraintLLM~\cite{shi-etal-2025-constraintllm} decompose constraints into retrievable units to assist formal reasoning. MIRROR~\cite{shi2026mirror} leverages hierarchical search over exemplar libraries to provide task-specific modeling and code references. LEAN-LLM-OPT~\cite{liang2026largescaleoptimizationmodelautoformulation} retrieves reference examples from a curated reference dataset and constructs structured modeling workflows to guide downstream formulation and code generation; AlphaOPT~\cite{kong2025alphaopt} builds experience library that distills structured insights from failed attempts.

\subsection{Skill Learning in LLM Agents} 
In contrast to task-specific experience augmentation, LLM agents equipped with skills offer a more general framework for cross-task knowledge accumulation and reuse. Skills preserve actionable task knowledge yet remain sufficiently abstract to support transfer across diverse problems. XSkill~\cite{jiang2026xskill} separates operational experiences from structural skills. EvoSkill~\cite{alzubi2026evoskill}, Memento-Skills~\cite{zhou2026mementoskills}, and AutoSkill~\cite{yang2026autoskill} emphasize automatic skill discovery, routing, and evolution. Trace2Skill~\cite{ni2026trace2skill} inductively consolidates trajectory-local lessons into unified, conflict-free, and transferable skill catalogs. MemSkill~\cite{zhang2026memskill} recasts memory operations themselves as learnable skills, optimizing the skill library through a closed loop of controller, executor, and designer.

\section{Methodology}
\label{mathodchapter}
\subsection{Preliminaries}
\textbf{Optimization Problem Archetype.} An optimization problem archetype $a\in\mathcal{A}$ refers to a canonical optimization structure abstracted from concrete application scenarios. Given an archetype $a$, we denote its natural-language instance as $x=g_u(a)$, where $u$ specifies the concrete scenario and $g_u$ represents the verbalization process that expresses the underlying optimization structure as a natural-language problem statement. Ideally, $x$ and its archetype $a$ should induce the same optimization model, i.e., $f_{\mathrm{model}}(a)=f_{\mathrm{model}}(g_u(a))$. It indicates that optimization modeling should be invariant to surface scenario and narrative variations, and problems that share the same archetype may differ substantially in scenario settings, narrative contexts, and natural-language surface forms, yet they typically share similar decision variables, objective functions, constraint patterns, and parameter configurations at the modeling level.

\textbf{Skills.} A skill $s\in\mathcal{S}$ is a reusable procedural knowledge document tailored to a class of optimization problem archetypes. Formally, $s=(m,w,p)$, where $m$ denotes skill metadata, including its name and brief description; $w$ denotes the modeling and solving workflow, covering decision variable, objective construction, constraint modeling, solver configuration, status checking, and result parsing; $p$ denotes common pitfalls, which summarize frequent error patterns in modeling and solving. A skill example is provided in Appendix~\ref{appendix_skillex}.

\subsection{An Overview of \textsc{OptSkills}}
We present \textsc{OptSkills}, an agent system for natural-language optimization tasks. As shown in Figure~\ref{fig:framework}, \textsc{OptSkills} proceeds in three phases. \textbf{Phase I: Skill Discovery via Cluster-Based Distillation} constructs the initial skill library $\mathcal{M}_0$ from the dataset. It first derives problem archetypes through ingredients extraction and problem editing, then clusters problems in this representation space, and distills cluster-based skills from modeling and solving trajectories within each cluster. \textbf{Phase II: Skill Learning} refines and expands the library on newly encountered problems. \textbf{Phase III: Test Evaluation} retrieves skills relevant to the target problem from $\mathcal{M}$ and injects them into the agent as procedural guidance for inference.

\subsection{Phase I: Skill Discovery via Cluster-Based Distillation}

\textbf{Problem Archetype Representation.} Given a problem $x_i$, the system first constructs an archetype embedding $e_i$ associated with its archetype $a_i$. This embedding captures the problem's modeling structure while reducing the influence of scenario-specific narratives. An extractor driven by an LLM outputs optimization ingredients $\kappa_i$ and an edited problem description $\tilde{x}_i$.

The ingredients $\kappa_i$ include the variables, constraints, and objective function at an abstract level. They are not copied from the original text, but are designed to ignore concrete scenarios, entity names, and narrative backgrounds as much as possible. The edited text $\tilde{x}_i$ replaces scenario-specific narratives and entity names with generic placeholders while preserving the original parameters, constraints, and numerical information. 

The archetype embeddings $e_i = \mathrm{Norm}(\alpha\cdot w_i + (1-\alpha)\cdot v_i)$ are calculated as the normalized weighted sum of $w_i$ and $v_i$, which are the embeddings of $\kappa_i$ and $\tilde{x}_i$, respectively. We employ Qwen3-Embedding-v3 as the embedding model, and the weight $\alpha \in [0,1]$ controls the relative contributions of $\kappa_i$ and $\tilde{x}_i$. Problems that share a similar archetype $a$ are expected to have high similarity between their archetype embeddings $e_i$, which provides the basis for clustering.

\textbf{Solver Portfolio Rollout.} For each problem, this module explores multiple different modeling approaches and solver configurations, and collects solution trajectories. The rationale is that different modeling paradigms and solver configurations may vary in their applicability and stability for the same optimization problems. Given $\tilde{x}_i$ and $\kappa_i$, an LLM-based solver selector chooses a subset $\mathcal{B}_i$ consisting of the top-$k$ most likely solvers from the candidate pool $\mathcal{B}$. For each solver $b \in \mathcal{B}_i$, the LLM first generates an optimization model $z_i^b$. It then iteratively generates executable solver code $c_{i,t}^b$ and receives execution observations $o_{i,t}^b$.

Assume that after $T_i^b$ iterations of code generation and solving, a feasible solution $\hat{y}_i^b$ is obtained. We define the modeling and solving trajectory under the specific combination of modeling paradigm and solver configuration $b$ as 
\begin{equation}
\tau_i^b = \bigl( z_i^b, c_{i,1}^b, o_{i,1}^b, c_{i,2}^b, o_{i,2}^b,\dots, c_{i,T_i^b}^b, \hat{y}_i^b \bigr).
\label{eq:trajectory_definition}
\end{equation}
The trajectory set is $\mathcal{T}_i=\{\tau_i^b\mid b\in\mathcal{B}_i\}$. In the trajectory set, a trajectory is labeled positive if $\hat{y}_i^b$ matches the ground-truth answer $y_i^{\ast}$; otherwise, it is labeled negative. This partitions $\mathcal{T}_i$ into positive trajectory set $\mathcal{T}_i^{+}$ and negative trajectory set $\mathcal{T}_i^{-}$.

\textbf{Trajectory Analysis.} Given $\mathcal{T}_i^{+}$ and $\mathcal{T}_i^{-}$, an LLM-based skill analyst is used to distill rollout trajectories into readable analyses. From positive trajectories, it extracts effective solver configurations and reusable modeling and solving procedures, summarized as standard operating procedures (SOPs). From negative trajectories, it identifies failure causes and summarizes them as common pitfalls:
\begin{equation}
q_i = A_{\eta}(\mathcal{T}_i^{+},\mathcal{T}_i^{-}) = (r_i,p_i).
\label{eq:trajectory_analysis}
\end{equation}

The skill analyst $A_{\eta}$ returns two components: reusable SOPs $r_i$ and the corresponding common pitfalls $p_i$. Each analysis $q_i$ serves as an input to skill distillation, contributing an atomic unit of distilled knowledge.

\textbf{Cluster-Based Skill Distillation.} After obtaining $e_i$, $\mathcal{T}_i$, and analysis $q_i$ for each problem, problems are clustered according to their archetype embeddings, and reusable skills are distilled at the cluster level. Specifically, it applies DBSCAN~\cite{dbscan} to embeddings, producing $\mathcal{C}=\mathrm{DBSCAN}(\{e_i\}_{i=1}^{N};\epsilon,m)$, where $\epsilon$ is the neighborhood radius and $m$ is the minimum number of samples required to form a core point. Since $e_i$ is normalized, cosine distance is used for clustering. Each cluster $C\in\mathcal{C}$ corresponds to a group of problems with a similar archetype.

For each cluster, an LLM-based skill builder aggregates the trajectory analysis of its member problems and distills a skill $s_C=B_{\eta}(\{q_i\mid i\in C\})$. The resulting set $\mathcal{M}_0=\{s_C\mid C\in\mathcal{C}\}$ forms the initial skill library. Cluster-based distillation aggregates diverse positive and negative analyses under the same problem archetype, yielding procedural knowledge that is less dependent on instance-specific artifacts and more suitable for reuse.

\subsection{Phase II: Skill Learning}
After constructing the initial skill library $\mathcal{M}_0$, \textsc{OptSkills} incrementally improves it on newly encountered problems. Given a new problem $x$, the system first applies the same ingredients extraction and problem editing steps as in Phase I. Based on these, an LLM-based skill selector identifies a candidate skill by comparing the problem against the names and descriptions of all skills in the current library. The selected candidate is then passed to an LLM-based judge, which determines whether the skill should be accepted for solving the problem. If a matching skill is accepted, the system refines it using evidence from the new solution trajectory. Otherwise, the problem is treated as a potential new archetype and may be used to expand the library when sufficient evidence is available.

\textbf{Skill Refinement.} When an existing skill $s^{\star}=(m^{\star},w^{\star},p^{\star})$ is selected, it is invoked as procedural guidance to solve the new problem $x$. The agent then generates a solution trajectory $\tau$, which is evaluated according to the correctness of the answer. A positive trajectory provides evidence for reinforcing the reusable SOPs in $w^{\star}$. A negative trajectory contributes failure evidence to the common pitfalls $p^{\star}$. The trajectory is distilled into an analysis $q$ by skill analyst $A_{\eta}$, and then used to update the selected skill $s^{\ast}$ into a new skill $s_{\text{refined}}$ by an LLM-based skill refiner.


The refiner updates the workflow component $w^{\star}$ using positive SOP evidence and augments the pitfall component $p^{\star}$ using negative evidence, while preserving the skill metadata and applicability boundary. The refined skill $s_{\text{refined}}$ replaces $s^{\star}$ in the library.

\textbf{Skill Expansion.} When no suitable skill is found, the new problem is treated as a candidate for a new archetype. \textsc{OptSkills} performs solver portfolio rollout to obtain a set of candidate trajectories $\mathcal{T}_x=\{{\tau_x^b\mid b\in\mathcal{B}_x}\}$. As in Phase I, the skill analyst summarizes them into trajectory analysis $q$.

To avoid adding unsupported procedural knowledge, \textsc{OptSkills} expands the library only when at least one positive trajectory is available. In that case, a skill builder constructs a new skill from the trajectory analysis as $s_{\mathrm{new}}=B_{\eta}(\{q\})$, and the library is updated to $\mathcal{M}_{t+1}=\mathcal{M}_{t}\cup\{s_{\mathrm{new}}\}$.

\subsection{Phase III: Test Evaluation}

During testing, \textsc{OptSkills} uses the learned skill library $\mathcal{M}^{\ast}$ as a fixed resource and does not perform any further updates. For each test problem $x$, the system first applies the same extractor as in previous phases to obtain the optimization ingredients $\kappa$ and the edited problem $\tilde{x}$. An LLM-based skill selector then identifies the most relevant skill from $\mathcal{M}^{\ast}$ according to $\kappa$ and $\tilde{x}$, relying on the skill names and descriptions as retrieval cues. The selected skill is used to guide the LLM agent during inference, and the prediction is finally evaluated against the ground-truth answer $y^{\ast}$.

\begin{table*}[t]
  \centering
  \scriptsize
  \setlength{\tabcolsep}{3pt}
  \caption{Comparison of the SA metric (Pass@1) across 5 benchmarks. \textsc{OptSkills-D} and \textsc{OptSkills-Q} indicate the \textsc{OptSkills} based on DeepSeek-V3.2 and Qwen3-235B-A22b-instruct-2507, respectively. \textbf{Bold} indicates the 1st, \underline{underline} indicates the 2nd, \uwave{wavy underline} indicates the 3rd.}
  \label{tab:general_sa_ranked}
  \resizebox{0.99\linewidth}{!}{
  \begin{tabular}{>{\centering\arraybackslash}m{2.25cm} | l | c c | c| c c c c c}
    \toprule
    Category & Models / Methods & \textbf{Macro-Avg.} & \textbf{Micro-Avg.} & \textbf{Rank} & \textbf{OptiBench} & \textbf{Mamo.C} &  \textbf{OptMATH} & \textbf{IndustryOR} & \textbf{ComplexOR}\\
    \midrule
    \multirow{4}{*}{\makecell{General Models}} 
       & GPT-5.4 & 51.71 & 57.82 & 6 & 69.09 & 47.39 & 46.39 & 29.00 & \uwave{66.67} \\
       & Gemini-3.1-Pro & {53.58} & 57.88 & {5} & {66.56} & 46.45 & \underline{54.22} & \underline{34.00} & \uwave{66.67} \\
       & Qwen3-235B & 42.63 & 52.36 & 7 & 63.80 & 41.71 & 39.76 & 29.00 & 38.89 \\
       & DeepSeek-V3.2 & 43.21 & 49.45 & 9 & 62.64 & 27.49 & 40.36 & 30.00 & 55.56 \\
    \midrule
    \multirow{4}{*}{\makecell{Agent-Based \\Methods}}

    & Chain-of-Experts & 38.84 & 52.36 & 8 & 62.31 & {51.66} & 35.54 & 28.00 & 16.67 \\
    & OptiMUS & 34.14 & 45.18 & 13 & 60.99 & 27.49 & 22.89 & 26.00 & 33.33 \\ 
    & ORMind & 36.69 & 47.37 & 11 & 61.82 & 34.60 & 21.69 & 32.00 & 33.33 \\
    & ORThought & 41.92 & 46.91 & 12 & 57.69 & 38.86 & 26.51 & 31.00 & 55.56 \\
    
    \midrule
    \multirow{2}{*}{\makecell{Experience-Enhanced\\Methods}}
    
    & LEAN-LLM-OPT & 41.23 & 49.27 & 10 & 62.15 & 43.13 & 22.89 & 28.0 & 50.0 \\ 
    & AlphaOPT & 50.92 & {58.55} & {4} & {70.25} & \underline{54.50} & 36.75 & 32.00 & 61.11 \\

    \midrule
    \multirow{3}{*}{\makecell{Skill-Based\\Methods}}
    & Trace2Skill & \underline{56.97} & \underline{63.46} & \underline{2} & \underline{75.21} & \uwave{54.03} & 49.40 & \underline{34.00} & \textbf{72.22} \\   

    & OptSkills-Q & \uwave{54.64} & \uwave{61.46} & \uwave{3} & \uwave{71.74} & 53.55 & \underline{54.22} & 27.00 & \uwave{66.67} \\   
    & OptSkills-D & \textbf{62.04} & \textbf{68.27} & \textbf{1} & \textbf{77.02} & \textbf{63.51} & \textbf{61.45} & \textbf{36.00} & \textbf{72.22} \\ 
    \bottomrule
  \end{tabular}
    }
\end{table*}

\section{Experiments}
\label{experiment}
\textbf{Datasets for Optimization Skills Construction.} To acquire problem instances encompassing diverse types and scenarios, we utilize the OptMATH-Train~\cite{lu2025optmathscalablebidirectionaldata} dataset as a candidate pool. We then perform uniform sampling across problem types and random sampling across scenarios, yielding 300 samples for \textsc{OptSkills} skill learning. The distribution across problem types is as follows: Linear Programming (75), Mixed-Integer Linear Programming (75), Integer Programming (75), Non-Linear Programming (45), and Second-order Cone Programming (30), spanning 20 distinct problem scenarios.

\textsc{OptSkills} first employs 150 instances for cluster-based skill distillation to derive an initial set of skills. Subsequently, \textsc{OptSkills} continues to learn on the remaining 150 samples. This phase serves to both augment the skill repository with new skills and refine existing skills. Detailed experiment setups are provided in Appendix~\ref{appendix_exp}.


\textbf{Evaluation Benchmarks.} For performance comparison, we adopt 5 benchmarks: OptiBench~\cite{yang2025optibench}, OptMATH-Bench~\cite{lu2025optmathscalablebidirectionaldata}, Mamo.C~\cite{huang2025llmsmathematicalmodelingbridging}, IndustryOR~\cite{Huang2025}, ComplexOR~\cite{xiao2024chainofexperts}. For Skill Learning evaluation, we employ the {hard split of NLCO Benchmark}~\cite{jiang2026reasoningcombinatorialconstrainedworld}. These benchmarks feature high problem difficulty and rich structural diversity. Detailed descriptions are provided in Appendix~\ref{appendix_benchmarks}.

\textbf{Baselines.} We benchmark our method against eleven distinct baselines in four categories. First, we include four general LLMs: GPT-5.4~\cite{openai2026gpt54}, Gemini-3.1-Pro~\cite{gemini31pro}, DeepSeek-V3.2~\cite{deepseekai2025deepseekv32pushingfrontieropen}, and Qwen3-235B~\cite{yang2025qwen3technicalreport}; these models directly generate solving code via prompting in a non-tool-use manner. Second, we compare with four LLM-based agent frameworks: CoE~\cite{xiao2024chainofexperts}, OptiMUS~\cite{pmlr-v235-ahmaditeshnizi24a}, ORMind~\cite{wang2025ormind}, and ORThought~\cite{yang2025orthought}. Third, we include two experience-enhanced agent-based approaches, AlphaOPT~\cite{kong2025alphaopt} and LEAN-LLM-OPT~\cite{liang2026largescaleoptimizationmodelautoformulation}, both of which support knowledge accumulation and reuse. Finally, we include Trace2Skill~\cite{ni2026trace2skill} as a strong skill-based baseline, since it also leverages distilled skills to enhance agent performance. Detailed reproduction settings are provided in Appendix~\ref{appendix_baselines}.

Through extensive experiments, we aim to answer the following research questions:

\textbf{Q1:} How does the optimization modeling and solving performance of \textsc{OptSkills} compare with baseline methods?

\textbf{Q2:} How do skills enhance the agent's capability in modeling and solving optimization problems?

\textbf{Q3:} What advantages does the clustering strategy offer for skill utilization?

\textbf{Q4:} How well does \textsc{OptSkills} generalize to in-distribution and out-of-distribution domains?

\textbf{Q5:} What are the characteristics of high-quality skills of \textsc{OptSkills}?

\textbf{Q6:} How effective is \textsc{OptSkills} in handling real-world and large-scale optimization problems?

\begin{table*}[t]
  \centering
  \scriptsize
  \setlength{\tabcolsep}{3pt}
  \caption{Ablation study of \textsc{OptSkills} on the SA metric across 5 benchmarks. The table evaluates the contribution of each module by incrementally adding them to the baseline model. A checkmark indicates that the corresponding module is enabled, while an empty box indicates that it is disabled.}
  \label{tab:ablation_sa}
  \resizebox{0.89\linewidth}{!}{
  \begin{tabular}{c c c c c | c c c c c}
    \toprule
    \makecell{Experimental Groups} 
    & \makecell{Tool-Call \\ Agent} 
    & \makecell{Skill \\Distillation} 
    & \makecell{Cluster-Based \\Distillation} 
    & \makecell{Skill \\Learning} 
    & \textbf{OptiBench} 
    & \textbf{Mamo.C} 
    & \textbf{OptMATH} 
    & \textbf{IndustryOR} 
    & \textbf{ComplexOR}\\
    \midrule

    Group 1 
    & \scalebox{1.5}{\CheckedBox} 
    & \scalebox{1.5}{\Square} 
    & \scalebox{1.5}{\Square} 
    & \scalebox{1.5}{\Square} 
    & 73.22 & 53.08 & 54.21 & 34.00 & 66.67 \\

    Group 2 
    & \scalebox{1.5}{\CheckedBox} 
    & \scalebox{1.5}{\CheckedBox} 
    & \scalebox{1.5}{\Square} 
    & \scalebox{1.5}{\Square} 
    & 69.75 & 56.40 & 59.64 & 32.00 & 66.67 \\

    Group 3 
    & \scalebox{1.5}{\CheckedBox} 
    & \scalebox{1.5}{\CheckedBox} 
    & \scalebox{1.5}{\CheckedBox} 
    & \scalebox{1.5}{\Square} 
    & 74.71 & 59.24 & 60.84 & 35.00 & {72.22} \\

    Group 4 
    & \scalebox{1.5}{\CheckedBox} 
    & \scalebox{1.5}{\CheckedBox} 
    & \scalebox{1.5}{\CheckedBox} 
    & \scalebox{1.5}{\CheckedBox} 
    & \textbf{77.02} & \textbf{63.51} & \textbf{61.45} & \textbf{36.00} & \textbf{72.22} \\

    \bottomrule
  \end{tabular}
  }
\end{table*}
\subsection{Main Results}
We evaluate all methods using Pass@1 solving accuracy (SA), and summarize the results in Table~\ref{tab:general_sa_ranked}.

\textbf{Comparison of \textsc{OptSkills} and Baseline Methods (To Q1).}
As summarized in Table~\ref{tab:general_sa_ranked}, \textsc{OptSkills} (using DeepSeek-V3.2 as the LLM backbone) consistently achieves state-of-the-art performance across almost all benchmarks, attaining a micro-average pass@1 accuracy of 68.27\% and demonstrating significant improvements over all baselines. To isolate the contribution of our system from the inherent capabilities of the underlying LLM backbone (DeepSeek-V3.2), we conduct an ablation study by using Qwen3-235B-A22b-instruct-2507 as the base model during inference. This configuration achieves a micro-average accuracy of 61.46\%, comparable to that of Trace2Skill, and outperforms the majority of baselines. These results suggest that the performance gains stem from \textsc{OptSkills} itself rather than from the choice of the backbone model.

Notably, \textsc{OptSkills} exhibits pronounced advantages on challenging benchmarks such as OptMATH and Mamo.C, where it substantially outperforms prompt-based baselines. Even on the relatively simple OptiBench benchmark, \textsc{OptSkills} maintains a significant accuracy margin.

\subsection{Ablation Study}
\label{section_ablation_study}
\textbf{The Contribution of Skills to Agent Performance (To Q2).} We conduct an empirical analysis of the cluster-based skill distillation process employed in Phase I. The Pass@1 accuracy of each condition across 5 benchmarks is reported in Table~\ref{tab:ablation_sa}. Although augmenting the baseline with per-sample skills (Group 2) yields modest gains on challenging benchmarks such as Mamo.C and OptMATH, the overall improvement remains marginal. This stems from functional overlap among skills. Moreover, per-sample extraction induces instance-specific biases, and skill selection relies on descriptions that primarily indicate applicability conditions rather than generalizable patterns. Consequently, individual per-sample skills exhibit limited transferability across structurally similar problems.

Moreover, we find that the system yields 129 distinct skills, of which 81 (62.8\%) are never invoked during inference in 5 benchmarks. However, these unselected skills are not inherently ineffective; when multiple semantically similar skills coexist, the model's likelihood of identifying a successful skill for a specific instance degrades markedly, compromising solution reliability.

\textbf{Enhancement of Skill Application via Clustering Strategies (To Q3).} In contrast, the clustered skill library (Group 3) exhibits substantial gains over the baselines. This shows that clustering significantly enhances the overall quality of the skill library, thereby facilitating robust generalization across problems that share the same archetype. 

In addition, we used cluster-based distillation to derive a set of 46 skills. Among them, only 7 skills remained unused throughout the inference process (15.2\%). In contrast, the non-clustering baseline exhibited a higher proportion of skills never being invoked (62.8\%). This indicates that the clustering-based distillation process effectively mitigates redundancy within the skill library, thereby enhancing the likelihood with which LLMs identify and retrieve the most appropriate skill for a given problem instance.

Since the clustering strategy has an impact on the quality of cluster-based skill generation, to validate the effectiveness of our proposed archetype-based clustering approach, we conduct a comparative analysis against the strategy that performs clustering directly on embeddings derived from raw problem text. Detailed results and discussions are provided in Appendix~\ref{appendix_knn}.

\textbf{Skill Learning on In-Distribution Data (To Q4-1).} As summarized in Group 4 of Table~\ref{tab:ablation_sa}, the proposed configuration demonstrates improvements in average accuracy across almost all benchmarks compared to Group 3. The exceptions are IndustryOR and ComplexOR, where the relatively high error rates~\cite{xiao2025asuvey} may weaken the statistical significance of the observed gains. Additionally, ComplexOR contains a comparatively small number of samples, which further limits the robustness of the comparison.


To investigate the growth dynamics of the skill library during skill learning and its impact on solving performance, we establish four checkpoints during \textsc{OptSkills}'s learning process, corresponding to skill library sizes of 46, 49, 53, and 56. We independently evaluate \textsc{OptSkills} configured with each checkpoint's skill library across multiple benchmarks. The macro-average solving accuracy exhibits an overall upward trend as the skill library grows from 46 to 56 skills (60.40\% $\rightarrow$ 60.88\% $\rightarrow$ 61.18\% $\rightarrow$ 62.04\%), while individual benchmarks show benchmark-specific fluctuations. The full scaling results are shown in Appendix~\ref{appendix_scaling}.

\begin{table}[t]
\centering
\caption{Performance comparison of \textsc{OptSkills} on the NLCO dataset. Before Learning denotes direct evaluation using the initial skill library obtained by cluster-based distillation from 150 samples in the previous dataset. After Learning denotes the setting where \textsc{OptSkills} further performs Skill Learning on our synthesized dataset.}
\label{tab:nlco_deepseek_v32}
\resizebox{0.8\linewidth}{!}{  
    \begin{tabular}{l| c}
    \toprule
    \textbf{Method} & \textbf{NLCO Accuracy (\%)} \\
    \midrule
    Tool-Call Agent & 62.60 \\
    Before Learning & 68.56 \\
    After Learning & \textbf{72.79} \\
    \bottomrule
    \end{tabular}
}
\end{table}

\subsection{Discussions}
\textbf{Skill Learning on Out-of-Distribution Data (To Q4-2).} To assess \textsc{OptSkills}'s generalization and adaptation ability under distribution shift, we first evaluate the initial skill library on the NLCO benchmark, an OOD combinatorial optimization benchmark. We then construct an OOD adaptation dataset, \textsc{NANO-CO}, via template-based synthesis. \textsc{NANO-CO} comprises combinatorial optimization problems whose problem-type distribution differs from the 300 samples for learning above. We use \textsc{NANO-CO} for skill learning and evaluate the adapted \textsc{OptSkills} again on NLCO.

As reported in Table~\ref{tab:nlco_deepseek_v32},  \textsc{OptSkills-D} with initial skill library (before learning) achieves a 5.96 percentage-point absolute improvement over Tool-Call Agent (68.56\% vs. 62.60\%), indicating that the initial skill library already exhibits nontrivial cross-domain transfer. 

Based on the 46 skills acquired in the initial skill library, we expanded the skill library with 47 new skills on \textsc{NANO-CO}, resulting in a total of 93 skills. We then evaluated \textsc{OptSkills} with the expanded skill library on the hard split of NLCO benchmark, achieving an accuracy of 72.79\%, a 10.19 percentage-point absolute gain over the baseline. This suggests that \textsc{OptSkills} can continuously expand its coverage of problem archetypes under shifted problem distributions.

Across all evaluated instances, successful trajectories exhibit a higher utilization rate of new skills compared to failed ones. Except for 9 failed samples, 644 of the 1,565 successfully solved samples employ new skills. In contrast, only 176 of the 576 failed samples utilized new skills, with the remainder relying exclusively on existing skills. To elucidate the contribution of newly learned skills, we plot the usage rate of old and new skills as a pie chart, which is provided in Appendix~\ref{appendix_usage}.

\textbf{OOD Performance at the Category Level (To Q4-3).} To further analyze the OOD adaptability of \textsc{OptSkills}, we categorize the NLCO samples into two groups based on whether their problem types were covered by \textsc{NANO-CO} and calculate the solving accuracy for each group separately. We divide the 43 problem types of NLCO into 22 categories with an exact or close counterpart in \textsc{NANO-CO} (Covered) and 21 without one (Uncovered) and compute the results of the two subsets.

\begin{table}[t]
\centering
\caption{Accuracy of \textsc{OptSkills} on two non-overlapping NLCO subsets, partitioned according to whether their problem types are covered by \textsc{NANO-CO}.}
\label{tab:nlco_two_group}
\resizebox{0.8\linewidth}{!}{  
    \begin{tabular}{l| c c c}
    \toprule
    \textbf{NLCO Subset} & \textbf{Before} & \textbf{After} & \textbf{Gain} \\
    \midrule
    Archetype-covered & 62.64 & 69.45 & \textbf{+6.81} \\
    Archetype-uncovered & \textbf{74.76} & \textbf{76.29} & +1.53 \\
    \bottomrule
    \end{tabular}
}
\end{table}

These results in Table~\ref{tab:nlco_two_group} suggest that the improvement is \textbf{not solely explained by the direct coverage of category}. As expected, the gain is substantially larger when the learned archetype is represented during adaptation (\textbf{+6.81}), demonstrating successful transfer across independently constructed datasets. At the same time, a consistent improvement remains on \textbf{archetype-uncovered categories (+1.53)}, indicating that the learned procedural knowledge also provides a degree of OOD adaptability.
Because these subsets differ in category composition and difficulty, we compare \textbf{within-subset improvements rather than their absolute accuracies}.

\textbf{Skill-level Performance (To Q5).} We group all instances across the benchmarks according to the skill used by the agent during problem solving. For each skill group, we compute the solving accuracy and compare it with the accuracy of the no-skill ablation on the same group of instances. We use the resulting accuracy difference ($\Delta \mathrm{Acc}=\mathrm{Acc}_{\mathrm{w/skill}}-\mathrm{Acc}_{\mathrm{w/o\ skill}}$) as an empirical measure of the contribution of each skill. To reduce the effect of small-sample fluctuations, we exclude groups containing fewer than 30 instances.

We then manually inspect the skills with the largest positive and negative accuracy gains. Specifically, the two best-performing skills, ``\texttt{fixed\_charge\_production\_planning}'' (usage count $N=45$, $\Delta \mathrm{Acc} = +11.11\%$) and ``\texttt{job\_shop\_scheduling\_with\_disjunctive \_constraints}'' ($N=47$, $\Delta \mathrm{Acc} = +8.51\%$), share two properties. First, they address structurally challenging problems and provide non-trivial modeling techniques, such as binary activation variables, disjunctive constraints, and Big-M formulations. Second, they offer multiple modeling and solving workflows, allowing the agent to adapt its strategy based on the specific instance and intermediate solver feedback. In contrast, the two lowest-performing skills, ``\texttt{multi \_commodity\_resource\_allocation\_with\_shared \_capacity}'' ($N=51$, $\Delta \mathrm{Acc} = -3.92\%$) and ``\texttt{integer\_resource\_allocation\_with\_demand \_satisfaction}'' ($N=66$, $\Delta \mathrm{Acc} = -4.55\%$), mainly describe relatively standard resource-allocation problems that the base agent can already solve well, and their workflows provide limited additional guidance.

\textbf{Solving Capability on High-Dimensional and Large-Scale Problems (To Q6).} To evaluate the capability of \textsc{OptSkills} on real-world and high-dimensional scenarios, we perform an evaluation on the challenging benchmark MIPLIB-NL~\cite{li2026constructing}, which is characterized by large-scale decision variables and complex block-structured constraints and requires high-order solution techniques. The results show that \textsc{OptSkills-D} achieves a solving accuracy of 26.91\% and exceeds the DeepSeek-V3.2-Thinking result reported in the MIPLIB-NL paper (22.38\%) by 4.53 percentage points. \textit{\textbf{Among the problems correctly solved by \textsc{OptSkills}, the dimensionality of decision variables reaches up to 87,482.}} This indicates that \textsc{OptSkills} maintains competitive performance even when confronted with large-scale, industrial problem instances.

\begin{table}[t]
    \centering
    \scriptsize
    \setlength{\tabcolsep}{4pt}
    \caption{
    Comparison of solving accuracy and computational cost among competitive methods. Token consumption is reported in millions (M).
    }
    \label{tab:cost_main}
    \resizebox{\columnwidth}{!}{
    \begin{tabular}{l|ccc}
        \toprule
        Method & Micro-Avg. (\%) & Inference & Learning \\
        \midrule
        AlphaOPT & 58.55 & 14.482M & 21.450M \\
        Trace2Skill & 63.46 & 40.269M & 12.518M \\
        OptSkills
        & \textbf{68.27}
        & 59.635M
        & 19.262M \\
        \bottomrule
    \end{tabular}
    }
\end{table}

\textbf{Computational Cost.}
Table~\ref{tab:cost_main} shows that the computational overhead of \textsc{OptSkills} is concentrated primarily at inference time. While its training cost remains comparable to those of AlphaOPT and
Trace2Skill, \textsc{OptSkills} incurs a higher inference cost while achieving the best overall solving accuracy. This cost profile suggests that the additional computation is mainly associated with the skill selection and multi-round tool-based solving at test-time. Consequently, the principal efficiency bottleneck of the current system lies in inference. 
The inference costs of \textsc{OptSkills} and the baselines, together with detailed component-level breakdowns of the learning and inference costs of \textsc{OptSkills}, are provided in Appendix~\ref{app:cost_analysis}.

\section{Conclusion}
\label{conclusion}
This paper proposes \textsc{OptSkills}, an agent system that clusters optimization problems from problem archetypes and distills reusable skills from diverse solving trajectories. Across a diverse collection of problems spanning multiple problem types and scenarios, these archetype-level skills improve solving performance and optimization generalization compared with various baselines. Extensive experiments demonstrate that \textsc{OptSkills} exhibits strong optimization generalization, enabling it to effectively tackle challenging and large-scale optimization problems. Moreover, it demonstrates strong generalization to out-of-distribution (OOD) problems and possesses the capacity for continuous expansion to new problem types and scenarios.

\section*{Limitations}
\label{limitation}

\textsc{OptSkills} relies on LLM-generated intermediate abstractions to construct problem archetype representations, including ingredients and edited problem descriptions. Although this design reduces the influence of surface narratives, errors in ingredient extraction or problem editing may propagate to clustering, skill selection, and downstream skill distillation. For example, missing a key constraint or incorrectly abstracting the objective may place a problem into an inappropriate cluster and lead to less reliable skill reuse. Future work should incorporate stronger verification mechanisms for archetype extraction, such as symbolic consistency checks. In addition, the quality of the learned skills library depends on the granularity of the clustering of archetypes. Over-fragmented clusters may produce redundant skills that increase the difficulty of skill selection, whereas over-merged clusters may combine different problem archetypes and yield ambiguous procedural guidance. This may partly explain why increasing the number of skills does not always lead to monotonic improvements across benchmarks. Although our sensitivity analysis suggests that the selected clustering setting provides a reasonable trade-off, developing adaptive clustering, skill merging, and skill pruning mechanisms remains an important direction for future work.

\section*{Acknowledgments}
\label{limitation}
This work is supported by the National Natural Science Foundation of China (No. 62476091), Ant Group, and the Shanghai Municipal Special Program for Basic Research on General AI Foundation Models (2025SHZDZX026D08).

\bibliography{custom}
\appendix
\section*{Appendix}

\begin{figure*}[t]
    \centering
    \includegraphics[width=0.8\textwidth]{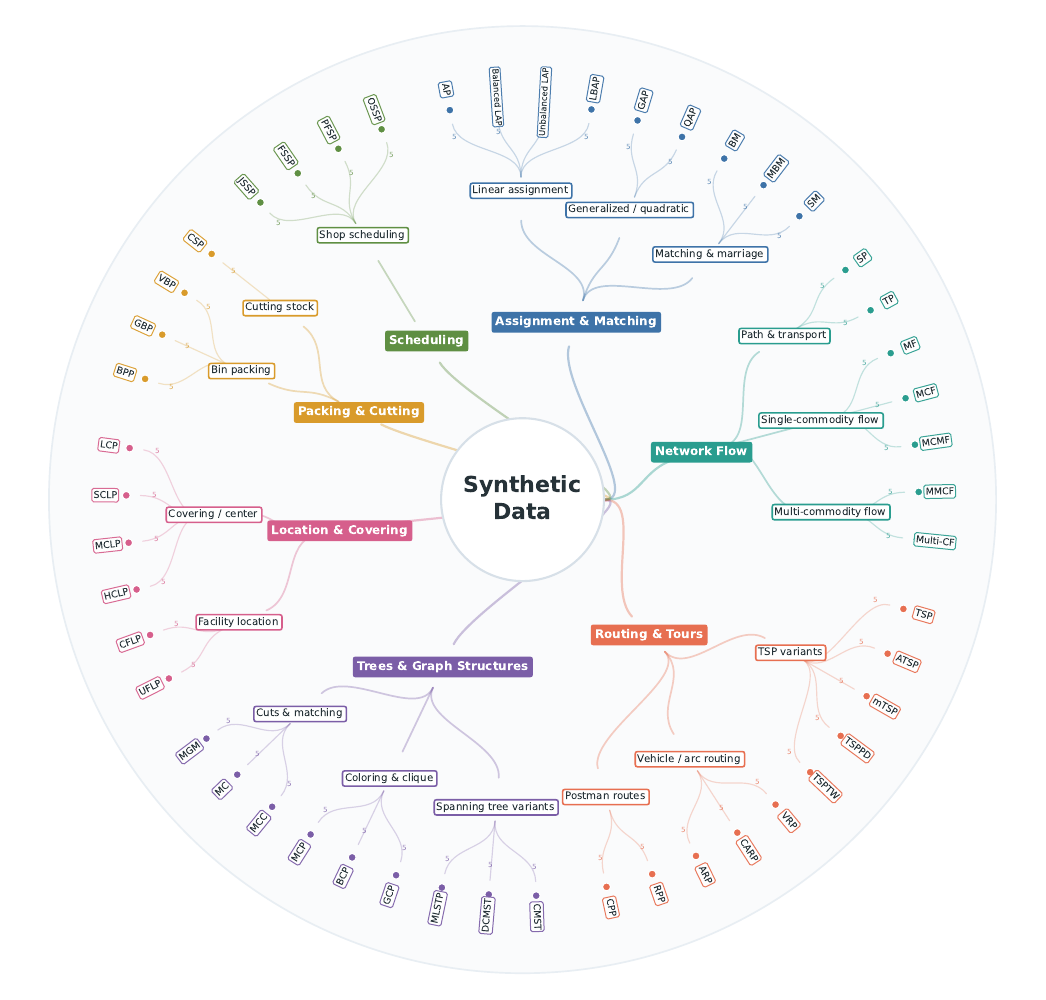}
    \caption{
    \textbf{Taxonomy of the synthetic data.} The dataset contains 49 combinatorial optimization problem types grouped into seven families. Each leaf node represents one problem type.
    }
    \label{fig:synthesis_data}
\end{figure*}

\begin{table*}[t]
    \centering
    \scriptsize
    \setlength{\tabcolsep}{3.5pt}
    \caption{Solver, execution-timeout, and tool-access settings for agent-based methods.}
    \label{tab:baseline_execution_settings}
    \resizebox{\textwidth}{!}{
    \begin{tabular}{l|cccccccc}
        \toprule
        \textbf{Setting}
        & \textbf{CoE}
        & \textbf{OptiMUS}
        & \textbf{ORMind}
        & \textbf{ORThought}
        & \textbf{LEAN-LLM-OPT}
        & \textbf{AlphaOPT}
        & \textbf{Trace2Skill}
        & \textbf{\textsc{OptSkills}} \\
        \midrule

        Solver backend
        & Gurobi
        & Gurobi
        & PuLP/CBC
        & Gurobi
        & Gurobi
        & Gurobi
        & \makecell{Skill-guided selection\\among available backends}
        & \makecell{Skill-guided selection\\among available backends} \\

        Execution timeout
        & 120 s
        & 120 s
        & 120 s
        & 120 s
        & 120 s
        & 400 s
        & \makecell{120 s per\\\texttt{run\_code} call}
        & \makecell{120 s per\\\texttt{run\_code} call} \\

        Tool access
        & None
        & None
        & None
        & None
        & \makecell{Experience\\retrieval}
        & \makecell{Experience\\retrieval}
        & \texttt{run\_code}
        & \texttt{run\_code} \\

        \bottomrule
    \end{tabular}
    }
\end{table*}

\section{Data Synthesis}
\label{appendix_datasynthesis}
To evaluate and enhance the generalizable modeling capability of \textsc{OptSkills} on combinatorial optimization problems, we constructed a collection of synthetic combinatorial optimization instances, named \textsc{NANO-CO}, as an adaptation set for skill learning before evaluation on the held-out NLCO benchmark. Figure~\ref{fig:synthesis_data} summarizes the taxonomy of this synthetic dataset, including its high-level problem families, subfamilies, and individual problem types.

We manually curated a set of common combinatorial optimization problems and their representative variants. Based on these problem categories, we further developed 49 seed code templates, with each seed template corresponding to one problem type.

All seed templates were manually inspected and verified using solvers. Particular attention was paid to potential implementation issues in the objective functions, constraints, decision-variable definitions, and parameter settings that could affect optimality, ensuring that each template can consistently generate instances with verifiable global optimal solutions. Then, while preserving the original mathematical structure, we used large language models to generate five new themes with realistic business contexts for each template, such as logistics scheduling, emergency department rostering, and semiconductor manufacturing. 

To ensure that thematic rewriting does not alter the original combinatorial optimization structure, we established a strict entity-mapping relationship for each new theme. For example, a ``machine'' in the original problem may be mapped to an ``operating room,'' while its modeling role, constraint relationships, and parameter semantics remain unchanged. For each themed instance, we further injected an independent random seed to generate the corresponding numerical parameters and invoked solvers to verify both feasibility and global optimality. This process yielded a total of 245 training instances.

\begin{figure}[t]
    \centering
    \includegraphics[width=\columnwidth]{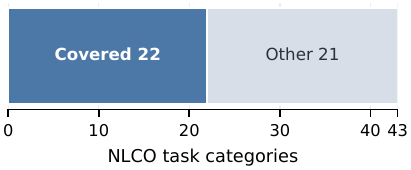}
\caption{
\textbf{NLCO task-category coverage} by \textsc{NANO-CO} exact or close counterparts.
}
\label{fig:nlco_coverage}
\end{figure}

To contextualize the relationship between \textsc{NANO-CO} and NLCO, we manually compared their problem categories. An NLCO task category is counted as covered if it has an exact or close counterpart in \textsc{NANO-CO}. As shown in Figure~\ref{fig:nlco_coverage}, \textsc{NANO-CO} covers 22 of the 43 NLCO task categories, indicating substantial overlap while still leaving room for NLCO-specific problem types. Regarding instance scale, the number of constraints in \textsc{NANO-CO} ranges from 1 to 5, with an average of 2.8, and the number of decision variables ranges from 9 to 310, with an average of 48.

\section{Experiment Setup}
\label{appendix_exp}
\subsection{Configuration of \textsc{OptSkills}}
Throughout all experiments, DeepSeek-V3.2 (non-thinking mode) serves as the uniform underlying LLM for all modules of \textsc{OptSkills}. The embedding model is Qwen3-Embedding-v3. We set the maximum agent-loop turns to 12 and set the temperature of the LLM backend to 0 across all stages.
When fusing the archetype embeddings, we set the parameter $\alpha$ to 0.55.
In the solver portfolio stage, the top-$k$ parameter is set to 3.
For DBSCAN clustering, the density parameter $\epsilon$ is set to 0.05, and the minimum number of samples $m$ is set to 1.
The random seed for training-data shuffling is set to 42.

\subsection{Experimental configuration of Baselines}

All agent-based methods use DeepSeek-V3.2 in non-thinking mode with temperature 0 as base model API call, receive the same benchmark inputs, and are evaluated under the same Pass@1 protocol. Unless otherwise specified, we follow the default settings of the original papers or released implementations.

Because the compared methods use different native inference workflows, we retain their original components and rounds rather than enforcing identical numbers of LLM calls, executions, or repair iterations with OptSkills. Table~\ref{tab:baseline_execution_settings} summarizes the main solver, timeout, and tool-access settings.

We retain the native inference and repair rounds of each method: CoE uses its three-round backward-reflection stage; OptiMUS permits up to three executions through its debugging loop; ORMind uses an adaptive 2--5 round repair budget; ORThought performs one Detection--Diagnosis--Repair pass; LEAN-LLM-OPT uses no additional repair loop; and AlphaOPT allows up to five debugging retries for runtime errors. Trace2Skill and \textsc{OptSkills} use a maximum end-to-end function-call loop budget of 12 turns.

For AlphaOPT and Trace2Skill, we make necessary adaptations to ensure compatibility with our experimental setting and evaluation protocol. Detailed modifications are provided in Appendix~\ref{appendix_alphaopt} and Appendix~\ref{appendix_trace2skill}.

\section{Benchmarks}
\label{appendix_benchmarks}
To ensure the reliability and consistency of our experimental results, we adopt the cleaned and corrected versions of OptiBench, Mamo.C, OptMATH, IndustryOR, and ComplexOR from LLMOPT \cite{JiangShu2025llmopt}, rather than the original datasets.

\subsection{OptiBench}
OptiBench is a large-scale benchmark designed to evaluate the end-to-end capability of large language models in solving optimization problems from complex inputs. It covers a diverse set of optimization settings, including linear programming, nonlinear optimization, and problems involving tabular data, reflecting practical scenarios where optimization tasks may be specified in heterogeneous formats. By incorporating nonlinear objectives or constraints as well as table-based problem information, OptiBench provides a comprehensive testbed for assessing whether models can accurately understand problem descriptions, formulate optimization models, and produce correct solutions across multiple forms of input.

\subsection{Mamo.C}
Mamo.C refers to the Complex\_LP subset of the Mamo benchmark, which is designed to evaluate the mathematical modeling ability of large language models through solver-based answer verification. It contains 211 undergraduate-level optimization problems integrating LP and MILP structures. In this benchmark, a model is required to translate a natural-language problem description into a solver-readable optimization formulation, and the generated formulation is evaluated by comparing the solver-produced optimal value with the ground-truth answer. Since Mamo.C involves more complex decision variables and constraints than the Easy\_LP subset, it provides a stricter testbed for evaluating optimization modeling capability.

\subsection{OptMATH}
OptMATH-Bench is a difficult optimization modeling benchmark derived from the OptMATH bidirectional data synthesis framework. It consists of hard instances selected through solver-based validation and rejection sampling, with problem descriptions that are substantially longer than those in NL4OPT~\cite{pmlr-v220-ramamonjison23a} and MAMO. The benchmark covers a broad range of optimization problem types, including LP, MILP, IP, NLP, and SOCP, providing a challenging testbed for evaluating whether large language models can formulate and solve complex long-context optimization problems.

\subsection{IndustryOR}
IndustryOR is an industrial benchmark introduced in the ORLM framework for evaluating large language models on practical operations research problems. Unlike benchmarks that mainly focus on textbook-style or homogeneous linear programming instances, IndustryOR is designed to reflect real-world optimization modeling scenarios across multiple industries. It contains 100 practical OR problems covering five types of optimization tasks, including linear programming, integer programming, mixed-integer programming, nonlinear programming, and other optimization problems, with instances organized into three difficulty levels. The benchmark evaluates models by execution accuracy, where a generated model is considered correct if the optimal value obtained by executing the solver code matches the ground-truth optimal value.

\subsection{ComplexOR}
ComplexOR is a benchmark introduced by Chain-of-Experts (CoE) for evaluating LLMs on complex operations research modeling and programming problems. Unlike elementary NL4Opt-style instances, ComplexOR focuses on more realistic and challenging OR problems whose descriptions may involve implicit constraints, domain-specific terminology, and long reasoning chains. Models are required not only to understand the natural-language problem statement, but also to construct an appropriate optimization model and generate executable solver code. The benchmark evaluates whether the generated program can pass test cases annotated by OR specialists, thereby assessing the model's end-to-end capability in complex OR problem formalization and solving.
\begin{figure*}[h]
    \centering
    \includegraphics[width=\textwidth]{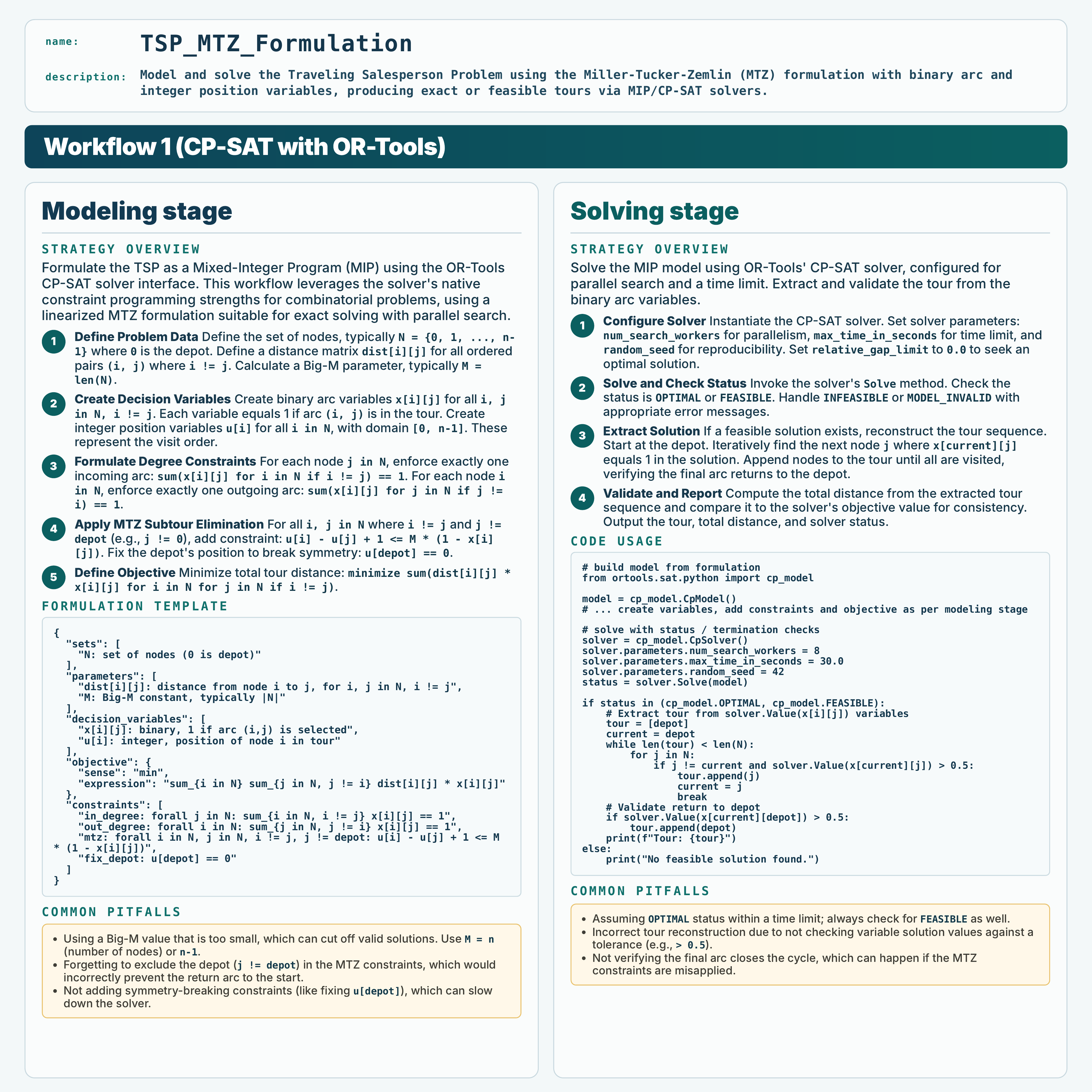}
    \caption{An example of the optimization skill.}
    \label{fig:skill_example}
\end{figure*}
\subsection{NLCO}
NLCO was originally proposed to evaluate the direct reasoning ability of large language models in natural-language combinatorial optimization. In its original protocol, given a natural-language decision-making scenario, the model directly outputs a discrete solution without writing code or invoking external solvers, and the output is evaluated for feasibility and optimality against problem constraints, objective functions, and solver-annotated reference solutions. Covering 43 classes of combinatorial optimization problems, NLCO spans a broad range of structures, including routing, scheduling, packing, assignment, facility location, and graph optimization.

In this paper, we use NLCO differently from its original tool-free evaluation setting. Rather than testing direct discrete-solution generation, we adopt NLCO as an out-of-distribution benchmark with natural-language problem statements, diverse combinatorial optimization structures, and verifiable reference answers. This enables us to examine whether \textsc{OptSkills} can generalize beyond the LP-, MILP-, and NLP-centered benchmarks used in our main experiments to a wider spectrum of combinatorial optimization archetypes.

Specifically, we evaluate \textsc{OptSkills} on NLCO’s most difficult Set-L, which contains 43 problem types with 50 instances each, totaling 2,150 test samples. None of these instances is used during \textsc{OptSkills}' skill learning or skill evolution process. During evaluation, \textsc{OptSkills} takes the original NLCO problem statement as input and obtains the final answer through skill retrieval, mathematical modeling, code generation, and execution. The output is then assessed using the ground truth provided by NLCO. Since Set-L enlarges the combinatorial search space and strengthens global constraint dependencies, it provides a rigorous test of \textsc{OptSkills}' generalization to unseen combinatorial optimization structures.

\subsection{MIPLIB-NL}
MIPLIB-NL is a high-dimensional and industrial-scale optimization modeling benchmark designed to expose the limitations of existing LLM-based optimization systems on large and difficult problems. It is constructed from real MILP instances in MIPLIB 2017 through a structure-aware reverse construction pipeline, yielding 223 verified natural-language-to-optimization instances with one-to-one correspondence to the original mathematical formulations. Since MIPLIB-NL targets problems whose scale can reach \(10^3\)--\(10^6\) variables and constraints, it provides a challenging testbed that is substantially closer to real industrial optimization than toy-scale benchmarks. 

In this work, we adopt MIPLIB-NL to assess the capability of \textsc{OptSkills} in handling high-dimensional optimization problems, especially whether the retrieved skills can still support reliable modeling and solving under large-scale, structurally complex settings.

\section{Skill Example}
\label{appendix_skillex}
An example of a skill is provided in Figure~\ref{fig:skill_example}. A skill contains multiple workflows. We show only one representative workflow here.

\begin{table*}[t]
    \centering
    \scriptsize
    \setlength{\tabcolsep}{6pt}
    \caption{Performance comparison with mean $\pm$ standard deviation over three independent runs.}
    \label{tab:statistical_analysis}
    \resizebox{\textwidth}{!}{
    \begin{tabular}{c|ccccc|cc}
        \toprule
        Method 
        & OptiBench 
        & Mamo.C 
        & OptMATH 
        & IndustryOR 
        & ComplexOR 
        & Macro-Avg. 
        & Micro-Avg. \\
        \midrule
        AlphaOPT 
        & 70.96 $\pm$ 0.63 
        & 54.03 $\pm$ 1.25 
        & 43.98 $\pm$ 6.38 
        & 31.67 $\pm$ 0.58 
        & 61.11 $\pm$ 0.00 
        & 52.35 $\pm$ 1.30 
        & 59.91 $\pm$ 1.25 \\
        
        Trace2Skill 
        & 76.14 $\pm$ 0.91 
        & 53.87 $\pm$ 0.72 
        & 49.00 $\pm$ 0.70 
        & 34.33 $\pm$ 0.58 
        & 74.07 $\pm$ 3.21 
        & 57.48 $\pm$ 0.65 
        & 63.94 $\pm$ 0.43 \\
        
        OptSkills 
        & 76.86 $\pm$ 0.29 
        & 62.09 $\pm$ 1.25 
        & 62.05 $\pm$ 1.04 
        & 36.00 $\pm$ 0.00 
        & 70.37 $\pm$ 3.21 
        & 61.47 $\pm$ 0.97 
        & 67.97 $\pm$ 0.45 \\
        \bottomrule
    \end{tabular}
    }
\end{table*}

\section{Baselines}
\label{appendix_baselines}

\subsection{Reproduction of Chain-of-Experts}
\label{appendix_coe}
Chain-of-Experts (CoE) is a multi-agent framework for solving operations research problems, where a conductor coordinates multiple specialized experts for terminology interpretation, mathematical modeling, programming, and code review. It constructs solutions through a forward expert-collaboration process and further refines them via backward reflection based on execution feedback. In our reproduction, we largely follow the default setting of the original implementation, including its expert configuration, conductor--reducer workflow, and forward--backward reflection mechanism. 

\subsection{Reproduction of OptiMUS}
\label{appendix_optimus}
OptiMUS-0.3 is a modular LLM-based agent system for formulating and solving linear and mixed-integer linear programming problems from natural-language descriptions. It decomposes the modeling process into a sequence of stages, including parameter extraction, clause formulation, code generation, code assembly, execution, and debugging, while maintaining an internal state and connection graph to organize the generated model components. In our reproduction, we use OptiMUS-0.3 without enabling its retrieval-augmented generation component. All other configurations follow the default implementation, including its modular modeling pipeline, error-correction and debugging mechanisms. 

\subsection{Reproduction of ORMind}
\label{appendix_ormind}
ORMind is a cognitive-inspired end-to-end reasoning framework for operations research problems. It follows a structured workflow that transforms natural-language requirements into mathematical models and executable solver code through several modules, including a Semantic Encoder, Formalization Thinking, an Executive Compiler, a System 2 Reasoner, and a Metacognitive Supervisor. A key feature of ORMind is its use of counterfactual reasoning to identify potential formulation or implementation errors and refine the generated solution. In our reproduction, we use the default setting of the original implementation, including its module configuration, memory-pool-based workflow, counterfactual reasoning mechanism, and syntax-error analysis.

\subsection{Reproduction of ORThought}
\label{appendix_orthought}
ORThought is a structured dual-agent framework for automated optimization modeling. It decomposes the solving process into two core agents: a Model Agent that performs problem understanding, mathematical modeling, and Gurobipy code generation, and a Solve Agent that executes the generated code and refines it through a Detection--Diagnosis--Repair workflow. In our reproduction, we follow the default setting of the original implementation, including its dual-agent architecture, prompt templates, structured reasoning process, and execution-based repair mechanism. No additional task-specific tuning is performed, and ORThought is evaluated under the same problem inputs and evaluation protocol as other baselines.

\subsection{Reproduction of LEAN-LLM-OPT}
\label{appendix_lean_llm_opt}
LEAN-LLM-OPT is a lightweight agentic workflow-construction framework for LLM-assisted optimization model auto-formulation. It uses upstream agents to identify the problem type and construct a structured workflow from reference examples, and a downstream model-generation agent follows this workflow to generate the final optimization formulation and executable solver code. The original framework relies on a reference dataset, Ref-Data, to support problem classification and workflow construction, and further incorporates CSV-based data-retrieval tools for large-scale instances with external datasets. 

In our reproduction, we enable its reference-based workflow construction using Ref-Data and use the \textsc{Without CSV} setting, since our evaluation inputs do not rely on external CSV files. Apart from this adaptation, we keep its default workflow-construction and model-generation procedure.

    

\begin{table*}[ht]
  \centering
  \scriptsize
  \setlength{\tabcolsep}{3pt}
  \caption{KNN retrieval performance comparing raw problem and archetype representations on NANO-CO and NLCO across multiple ranking metrics.}
  \label{tab:knn_result}
  \resizebox{0.88\linewidth}{!}{
  \begin{tabular}{c c | c c c c c c c}
    \toprule
    \textbf{Dataset} & \textbf{Representation} &
    \textbf{Hit@1} & \textbf{Hit@3} & \textbf{Hit@5} &
    \textbf{Precision@5} & \textbf{Recall@5} &
    \textbf{MRR} & \textbf{MAP@5} \\
    \midrule

    \multirow{2}{*}{NANO-CO}
    & Raw Problem
    & 0.416 & 0.690 & 0.808 & 0.344 & 0.430 & 0.582 & 0.318 \\

    & Archetype
    & 0.820 & 0.939 & 0.971 & 0.596 & 0.745 & 0.885 & 0.679 \\

    \midrule

    \multirow{2}{*}{\textcolor{black}{NLCO}}
    & \textcolor{black}{Raw Problem}
    & \textcolor{black}{0.558}
    & \textcolor{black}{0.805}
    & \textcolor{black}{0.880}
    & \textcolor{black}{0.503}
    & \textcolor{black}{0.051}
    & \textcolor{black}{0.697}
    & \textcolor{black}{0.418} \\

    & \textcolor{black}{Archetype}
    & \textcolor{black}{0.874}
    & \textcolor{black}{0.960}
    & \textcolor{black}{0.979}
    & \textcolor{black}{0.847}
    & \textcolor{black}{0.086}
    & \textcolor{black}{0.919}
    & \textcolor{black}{0.811} \\

    \bottomrule
  \end{tabular}
  }
\end{table*}

\subsection{Reproduction of AlphaOPT}
\label{appendix_alphaopt}
AlphaOPT is an experience-learning framework for natural-language optimization modeling. Instead of updating model parameters, it constructs a self-improving experience library of solver-verified modeling insights. Each insight is represented with explicit applicability information, including its taxonomy, condition, explanation, and example, so that it can be retrieved and reused for later optimization tasks. AlphaOPT operates through two stages: Library Learning extracts structured insights from failed attempts and organizes them in a hierarchical taxonomy, while Library Evolution refines the applicability conditions of stored insights using aggregate evidence across tasks to reduce over- and under-generalization.

In its original experimental setup, AlphaOPT performs a train-test split on the benchmark. To ensure a fair comparison in our reproduction of AlphaOPT, we use the same training set as our method and expand its experience library from scratch.

\subsection{Reproduction of Trace2Skill}
\label{appendix_trace2skill}
Trace2Skill is an automatic skill construction and adaptation framework for LLM agents. It distills reusable skills from execution trajectories by first collecting a pool of agent traces, then using parallel success and error analysts to propose trajectory-level skill patches, and finally consolidating these patches into a unified, conflict-free skill directory through hierarchical merging. The resulting skill is used as a declarative artifact without parameter updates or test-time retrieval.

\textcolor{black}{For Trace2Skill, we modify part of its prompts to adapt the method to the optimization problem setting and use the same training set as \textsc{OptSkills}.}

\section{Additional Experiments}
\subsection{Statistical Stability of Results}
\label{appendix_statistics}
\textcolor{black}{
To assess the stability of the main comparison, we repeat the evaluation of \textsc{OptSkills} and the two strongest competitive baselines, Trace2Skill and AlphaOPT, over three independent runs. 
Table~\ref{tab:statistical_analysis} reports the mean and standard deviation of Pass@1 across all five benchmarks, together with the macro- and micro-averaged results.
}

\textcolor{black}{
Across repeated runs, \textsc{OptSkills} preserves a clear overall advantage over both baselines. 
It achieves the highest mean accuracy on four of the five benchmarks and obtains the best Macro-Avg.\ ($61.47 \pm 0.97$) and Micro-Avg.\ ($67.97 \pm 0.45$). 
In particular, its Micro-Avg.\ exceeds Trace2Skill by 4.03 percentage points and AlphaOPT by 8.06 percentage points. 
OptSkills also exhibits low run-to-run variability on most benchmarks, with standard deviations no larger than 1.25 except on ComplexOR, where the small evaluation set leads to larger fluctuations. 
Overall, these results show that the performance gains of \textsc{OptSkills} are stable across repeated evaluations and extend across multiple benchmarks rather than being driven by an isolated result.
}


\begin{table}[t]
    \centering
    \color{black}
    \small
    \setlength{\tabcolsep}{4pt}
    \caption{Manual evaluation of the archetype extractor on 20 randomly sampled training problems. ``Score-2 Majority'' denotes the percentage of problems for which at least two of the three annotators assign a score of 2. Pairwise agreement denotes the mean exact agreement rate across the three annotator pairs.}
    \label{tab:extractor_manual_eval}
    \resizebox{\columnwidth}{!}{
    \begin{tabular}{l|ccc}
        \toprule
        \textbf{Evaluation Dimension}
        & \textbf{Mean Score (/2)}
        & \textbf{Score-2 Majority}
        & \textbf{Pairwise Agreement} \\
        \midrule
        Constraint Preservation & 1.82 & 90\% & 83.3\% \\
        Semantic Preservation   & 1.93 & 95\% & 96.7\% \\
        Ingredient Specificity  & 1.90 & 95\% & 83.3\% \\
        \bottomrule
    \end{tabular}
    }
\end{table}

\subsection{KNN-Based Validation of Embedding-Space Clusterability}
\label{appendix_knn}

To further validate the \textcolor{black}{discriminative capability of archetype representations} and the validity of using archetypes as the fundamental unit of abstraction, we conduct a K-Nearest Neighbor (KNN) validation. We establish a baseline condition in which raw problem texts are directly embedded to obtain representations $\tilde{e}$, and we compare these against our proposed archetype embeddings $e$. This experiment assesses the extent to which raw-text embeddings exhibit clusterable structure according to structural similarity rather than superficial semantic similarity in the embedding space. \textcolor{black}{We conduct this validation on both \textsc{NANO-CO} and NLCO to examine whether the representation advantage extends beyond the synthetic \textsc{NANO-CO} data to a real benchmark.}

We compute the following five metrics: \textbf{(1) Hit Rate (HR@k)}: Whether the top-$k$ neighbors contain at least one sample from the same archetype group. \textbf{(2) Precision@k}: The proportion of same-archetype samples among the top-$k$ neighbors. \textbf{(3) Recall@k}: The proportion of same-archetype samples retrieved among all members of that archetype group. \textbf{(4) Mean Average Precision (MAP)}: The mean of average precision scores across all queries. \textbf{(5) Mean Reciprocal Rank (MRR)}: The mean reciprocal rank of the first same-archetype neighbor.

All metrics range from 0 to 1, with higher values indicating better neighborhood consistency. As shown in Table~\ref{tab:knn_result}, \textcolor{black}{archetype embeddings outperform raw-text embeddings on both \textsc{NANO-CO} and NLCO. On NANO-CO, Hit@1 increases from 0.416 to 0.820 and MAP@5 from 0.318 to 0.679. On NLCO, Hit@1 increases from 0.558 to 0.874 and MAP@5 from 0.418 to 0.811. This suggests that archetype representations better preserve optimization structure and that the effect is not limited to synthetic data.} This corroborates the validity of using archetypes as the core abstraction unit for problem representation and skill organization.

\subsection{Manual Evaluation of Archetype Extraction}
\label{appendix_extraction_quality}

\textcolor{black}{
Since the archetype representation relies on LLM-generated optimization ingredients and edited problem descriptions, we conduct a manual evaluation to assess the reliability of this intermediate extraction stage. We randomly sample 20 problems from the 300 training instances used for skill construction and process them using the same extractor configuration as in the main experiments. For each problem, the annotators are provided with the original problem, the extracted optimization ingredients $\kappa_i$, and the edited problem $\tilde{x}_i$. Three annotators independently evaluate all 20 extraction results while being blinded to downstream solving trajectories, clustering results, and each other's ratings.
}

\textcolor{black}{
The annotation focuses on three dimensions related to whether the extractor preserves the underlying optimization structure:
\textbf{(1) Constraint Preservation}, whether the constraints in the original problem are completely preserved after extraction;
\textbf{(2) Semantic Preservation}, whether the edited problem retains the original optimization semantics without introducing or removing relevant information; and
\textbf{(3) Ingredient Specificity}, whether the extracted optimization ingredients remain sufficiently informative to characterize the problem structure rather than becoming overly generic.
Each dimension is rated on a three-level scale from 0 to 2, where 0 indicates a major error, 1 indicates minor information loss while preserving the core optimization structure, and 2 indicates no observable error.
}

\textcolor{black}{As shown in Table~\ref{tab:extractor_manual_eval}, the extractor achieves mean scores above 1.8 out of 2 across all three dimensions. In 90--95\% of the sampled problems, at least two annotators assign a score of 2. The average pairwise exact agreement across the three dimensions is 87.8\%, indicating substantial consistency among the annotators in assessing extraction quality.} \textcolor{black}{
Overall, the manual evaluation suggests that the extractor generally preserves the optimization structure and semantic information required for constructing archetype representations. However, the study is based on a limited sample of 20 problems and is intended as a reliability check rather than a comprehensive evaluation of extraction quality. Extraction errors may still occur and propagate to downstream clustering and skill retrieval. As a complementary evaluation of the resulting representation, Appendix~\ref{appendix_knn} compares archetype representations with raw problem-text embeddings on both \textsc{NANO-CO} and NLCO.
}



\begin{figure}[t]
    \centering
    \begin{subfigure}{0.48\textwidth}
        \centering
        \includegraphics[width=\columnwidth]{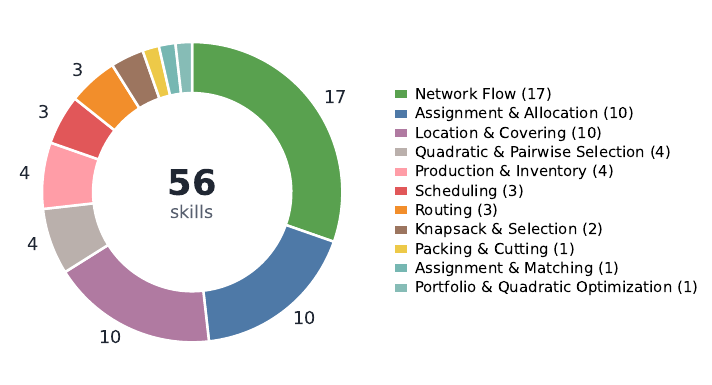}
        \caption{\textsc{OptMATH-Train}.}
        \label{fig:5a}
    \end{subfigure}
    \hfill
    \begin{subfigure}{0.48\textwidth}
        \centering
        \includegraphics[width=\columnwidth]{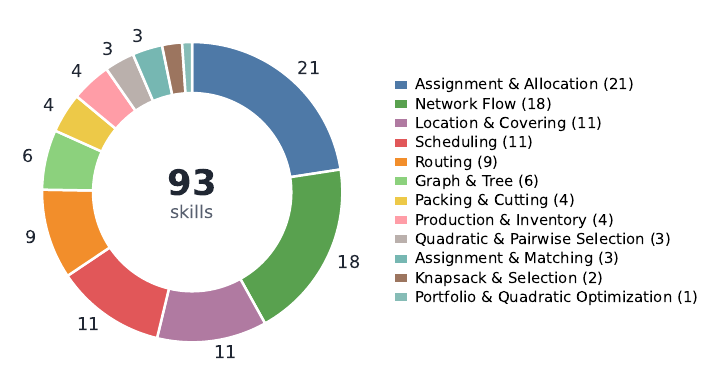}
        \caption{\textsc{NANO-CO}.}
        \label{fig:5b}
    \end{subfigure}
    \caption{
    \textbf{Skill coverage on problem family.} Top: coverage of the skill library learned from \textsc{OptMATH-Train}. Bottom: coverage after learning on \textsc{NANO-CO}.}
    \label{fig:coverage}
\end{figure}

\begin{figure*}[t]
    \centering

    \begin{subfigure}[t]{0.32\textwidth}
        \centering
        \includegraphics[width=\linewidth]
        {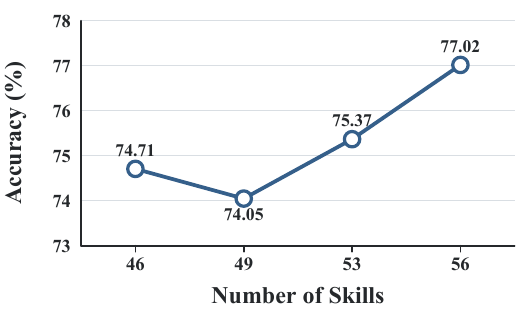}
        \caption{OptiBench}
    \end{subfigure}
    \hfill
    \begin{subfigure}[t]{0.32\textwidth}
        \centering
        \includegraphics[width=\linewidth]
        {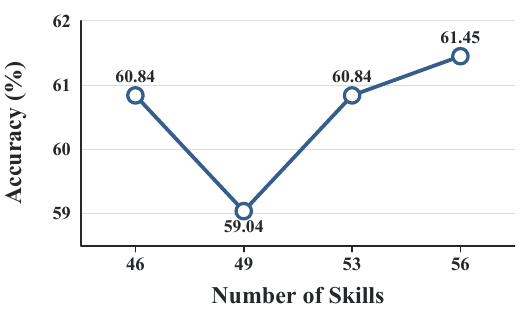}
        \caption{OptMATH}
    \end{subfigure}
    \hfill
    \begin{subfigure}[t]{0.32\textwidth}
        \centering
        \includegraphics[width=\linewidth]
        {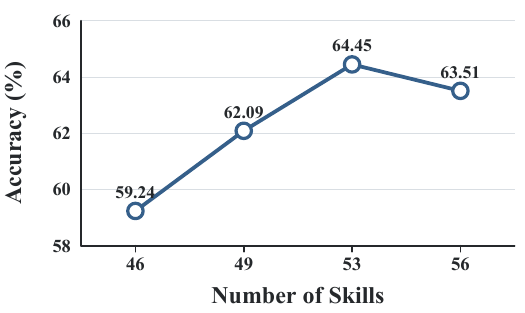}
        \caption{Mamo.C}
    \end{subfigure}

    \vspace{0.4em}

    \begin{subfigure}[t]{0.32\textwidth}
        \centering
        \includegraphics[width=\linewidth]
        {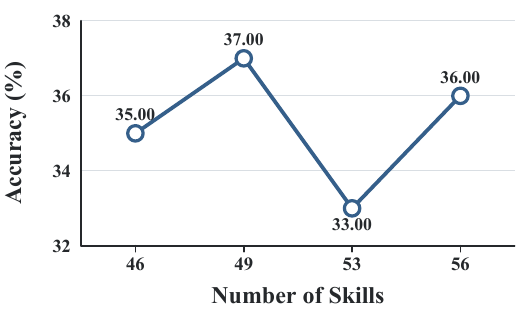}
        \caption{IndustryOR}
    \end{subfigure}
    \hfill
    \begin{subfigure}[t]{0.32\textwidth}
        \centering
        \includegraphics[width=\linewidth]
        {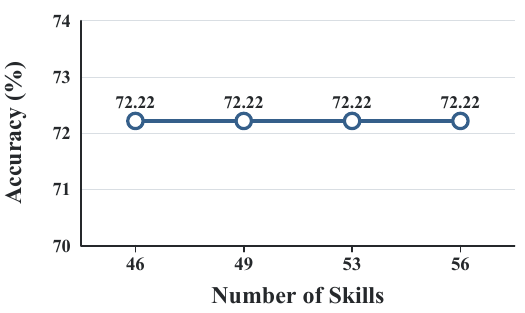}
        \caption{ComplexOR}
    \end{subfigure}
    \hfill
    \begin{subfigure}[t]{0.32\textwidth}
        \centering
        \includegraphics[width=\linewidth]
        {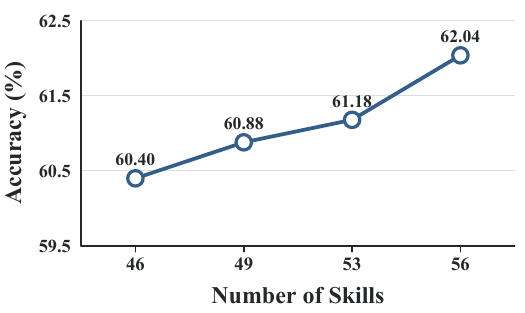}
        \caption{Macro-average}
    \end{subfigure}

    \caption{
        \textcolor{black}{Solving accuracy of \textsc{OptSkills}, based on DeepSeek-V3.2,
        across different benchmarks as the number of skills in the skill
        library increases. Each panel uses a benchmark-specific vertical
        axis range to make checkpoint-level performance changes visible.}
    }
    \label{fig:scaling_analysis}
\end{figure*}

\subsection{Analysis of Skill Coverage}
\label{appendix_skillcovery}
Figure~\ref{fig:coverage} visualizes the archetype-level coverage of the learned skill library. The skill library learned from \textsc{OptMATH-Train} contains 56 skills and already spans 11 optimization categories, including Network Flow, Assignment \& Allocation, Location \& Covering, Quadratic \& Pairwise Selection, Production \& Inventory, Scheduling, Routing, Knapsack \& Selection, and Packing \& Cutting. This suggests that the initial library captures a broad set of reusable modeling and solving patterns rather than being dominated by a single problem family.

Before evaluating on the OOD \textsc{NLCO} benchmark, we perform continued skill learning on our synthesized \textsc{NANO-CO} data. After this stage, the skill library expands to 93 skills and shows stronger coverage of combinatorial optimization structures, especially Assignment \& Allocation, Network Flow, Location \& Covering, Scheduling, Routing, Graph \& Tree, and Packing \& Cutting. These categories are highly aligned with the structural characteristics of NLCO.

This analysis confirms that \textsc{OptSkills} has strong skill coverage: the initial library provides broad in-domain archetype coverage, while continued learning on \textsc{NANO-CO} further extends the library toward OOD combinatorial structures. Since no skills are learned from \textsc{NLCO} itself, the expanded coverage supports the ability of \textsc{OptSkills} to transfer newly acquired skills to unseen problem distributions.

\subsection{The Scaling Exploration of Skills}
\label{appendix_scaling}
\textcolor{black}{
We investigate how the growth of the skill library affects solving accuracy across different benchmarks. 
As shown in Fig~\ref{fig:scaling_analysis}, we evaluate \textsc{OptSkills} with four skill-library sizes (46, 49, 53, and 56 skills) on five optimization benchmarks. 
The results show an overall upward trend in macro-average performance as more skills are acquired, increasing from 60.40\% to 62.04\%, although the improvement is not strictly monotonic for every benchmark.}

\textcolor{black}{Specifically, OptiBench, OptMATH, and Mamo.C exhibit an overall increasing trend as the skill library expands. 
OptiBench improves from 74.71\% to 77.02\%, while Mamo.C increases from 59.24\% to 63.51\%. 
OptMATH also achieves higher final performance with a larger skill library (60.84\% $\rightarrow$ 61.45\%), despite an intermediate decrease at the 49-skill checkpoint. 
In contrast, IndustryOR shows fluctuations across different checkpoints (35.00\%, 37.00\%, 33.00\%, and 36.00\%), and ComplexOR remains unchanged at 72.22\% throughout the scaling process.} \textcolor{black}{No clear patterns are observed for IndustryOR and ComplexOR, which can be attributed to two main factors: (1) these benchmarks are more challenging, as they primarily consist of complex MIP problems derived from real-world scenarios, so additional skills may not translate into consistent performance gains; and (2) according to the analysis in the MIPLIB-NL~\cite{li2026constructing}, the annotation quality of samples in these two benchmarks is imperfect, which may introduce additional noise into skill evaluation.}


\begin{figure}[t!]
    \begin{subfigure}[t]{0.48\linewidth}
    \includegraphics[width=0.95\linewidth]{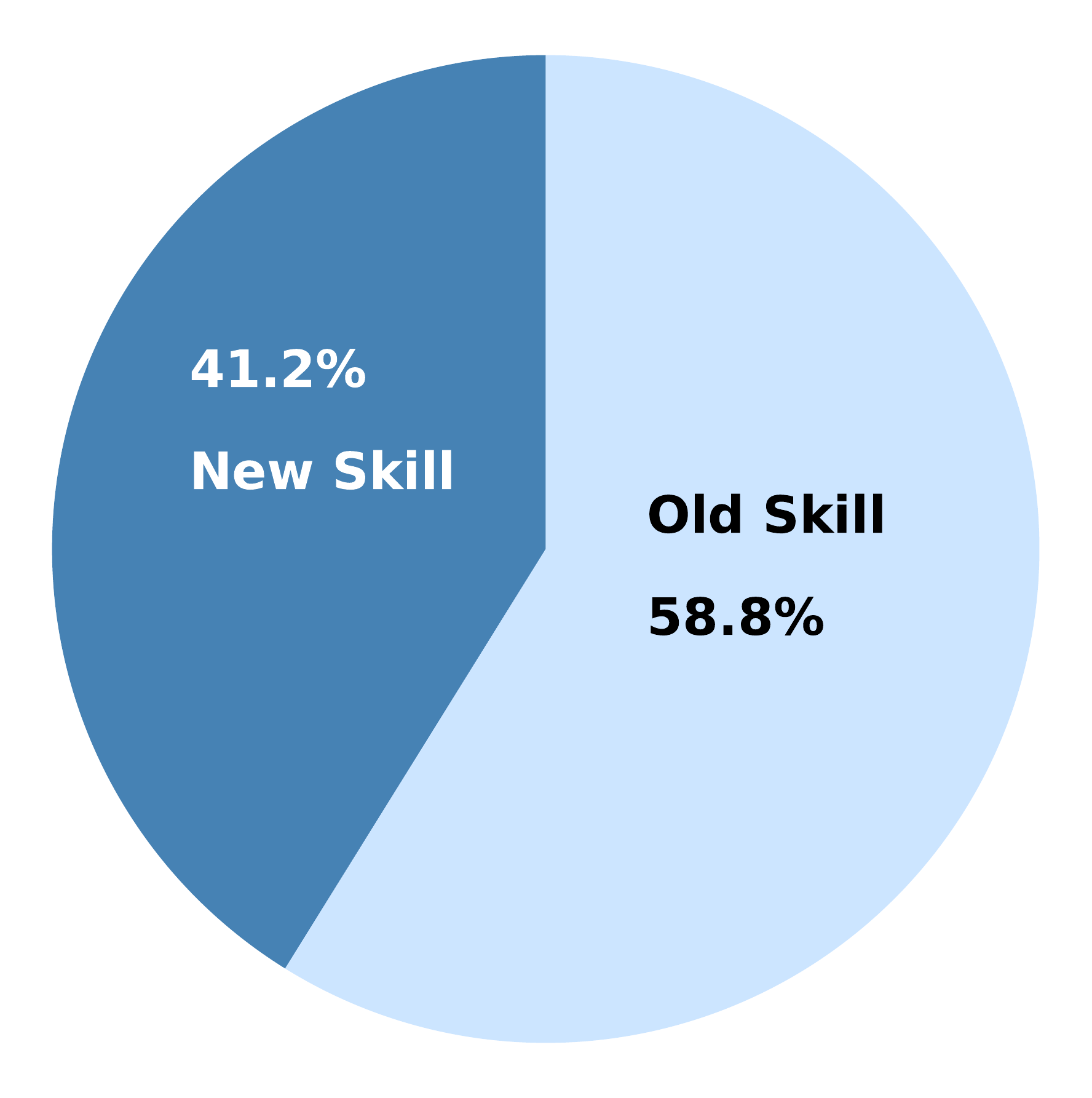}
        \caption{The skills usage of the successful samples.}
  \label{subfug_c}
  \end{subfigure}
  \hfill
    \begin{subfigure}[t]{0.48\linewidth}
    \includegraphics[width=0.95\linewidth]{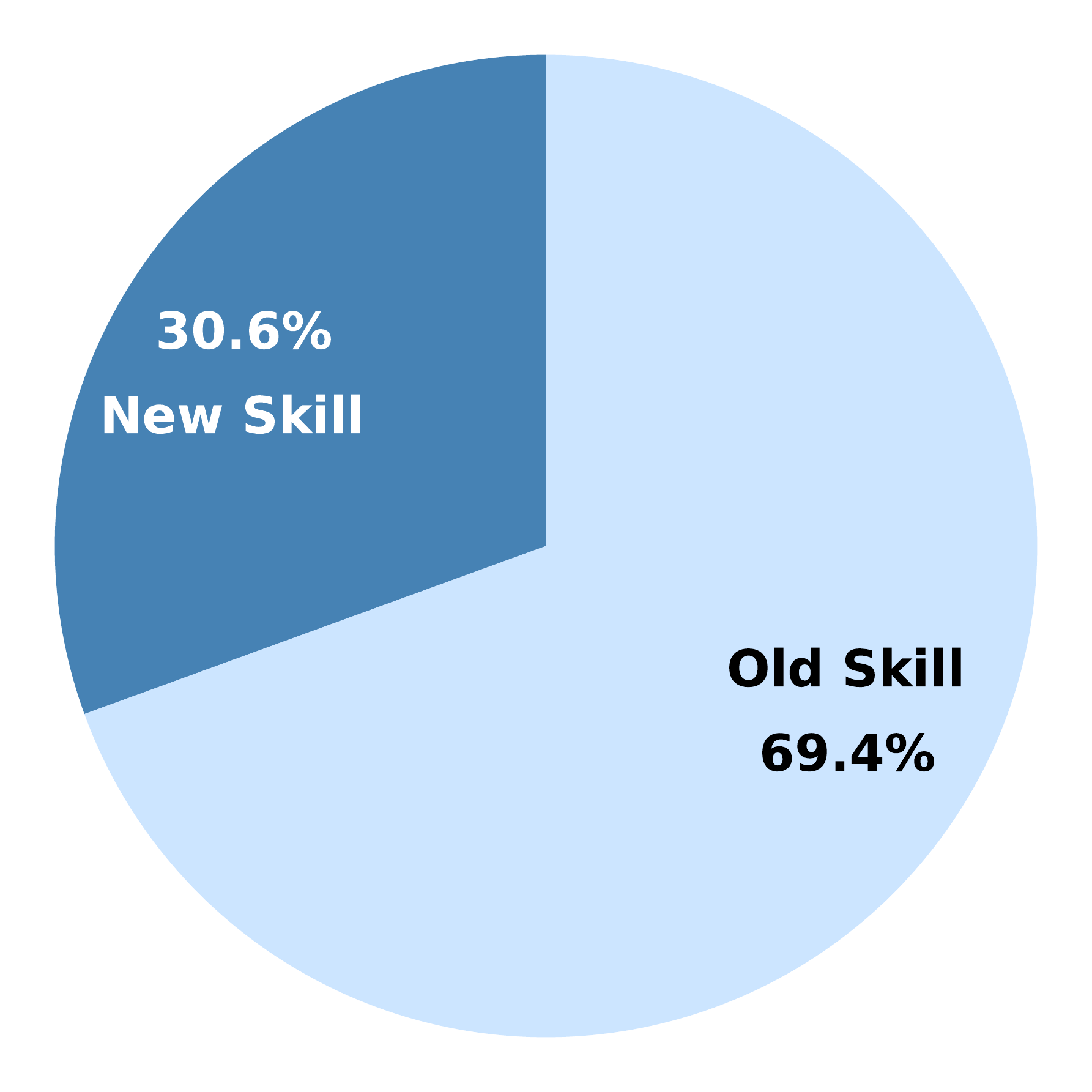}
        \caption{The skills usage of the failed samples.}
  \label{subfug_inc}
  \end{subfigure}
  \caption{The utilization of newly acquired and existing skills across the subsets of correctly solved and failed instances within the NLCO benchmark.}
  \label{fig:skillpie}
\end{figure}

\subsection{The Usage Statistics of New and Old Skills on NLCO Benchmark}
\label{appendix_usage}
On the \textsc{NANO-CO} dataset, \textsc{OptSkills} expands the initial set of 46 existing skills by acquiring 47 new ones. Figure~\ref{fig:skillpie} illustrates the utilization patterns of these old and new skills for both successfully solved and failed instances on the NLCO benchmark. The results demonstrate that the newly acquired skills contribute substantially to performance on this benchmark.

\subsection{Parameter Sensitivity Analysis}
\label{appendix_dbscansense}

\textbf{DBSCAN Clustering Diagnostics.} We evaluate the sensitivity of the archetype construction step to DBSCAN hyperparameters on \textsc{NANO-CO}. We fix the archetype embedding used in the main experiments and vary the DBSCAN radius $\epsilon$ from 0.01 to 0.15, together with $\texttt{min\_samples} \in \{1,2,3\}$. The ground-truth problem types are used only for post-hoc diagnostics and are not used during clustering.

\begin{figure}[ht!]
    \centering
    \begin{subfigure}[t]{\linewidth}
        \centering
        \includegraphics[width=\linewidth]{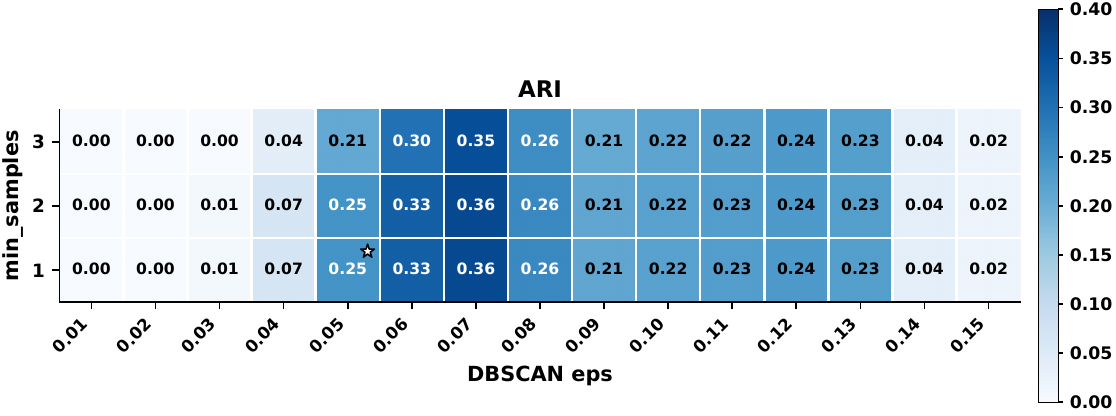}
        \caption{ARI}
        \label{fig:dbscan-sensitivity-ari}
    \end{subfigure}

    \begin{subfigure}[t]{\linewidth}
        \centering
        \includegraphics[width=\linewidth]{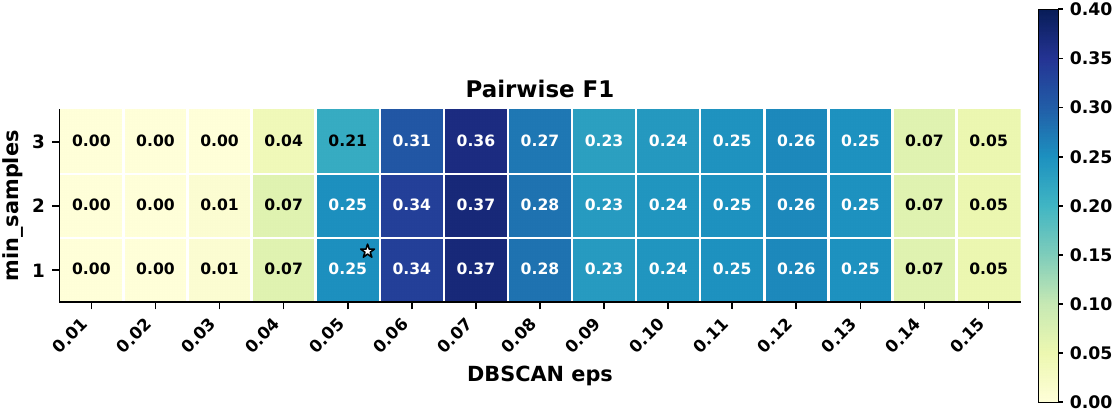}
        \caption{Pairwise F1}
        \label{fig:dbscan-sensitivity-f1}
    \end{subfigure}

    \caption{DBSCAN sensitivity analysis on \textsc{NANO-CO} archetype embeddings. The star marks the setting used in our main experiments.}
    \label{fig:dbscan-sensitivity}
\end{figure}

Figure~\ref{fig:dbscan-sensitivity} reports ARI and Pairwise F1 under different hyperparameter settings. Very small $\epsilon$ values lead to over-fragmented clusters and near-zero agreement with the problem-type labels. The clustering quality improves in a moderate range of $\epsilon$, with the best diagnostic scores around $\epsilon=0.07$. Although $\epsilon=0.07$ achieves the best post-hoc clustering diagnostic score, we use $\epsilon=0.05$ as a more conservative label-free setting to reduce the risk of merging structurally different archetypes.

\begin{table}[t!]
    \centering
    \color{black}
    \scriptsize
    \setlength{\tabcolsep}{3pt}
    \caption{Downstream sensitivity to the DBSCAN radius $\epsilon$ on Mamo.C.}
    \label{tab:dbscan_downstream_sensitivity}
    \resizebox{\linewidth}{!}{
    \begin{tabular}{c| c c c c c c c c c}
        \toprule
        $\epsilon$
        & 0.01 & 0.03 & 0.05 & 0.07 & 0.09 & 0.11 & 0.13 & 0.15 & 0.20 \\
        \midrule
        SA (\%)
        & 59.24 & 54.50 & 63.51 & 63.51 & 63.51 & 65.40 & 63.51 & 63.98 & 54.50 \\
        \bottomrule
    \end{tabular}
    }
\end{table}

\begin{table}[!t]
    \centering
    \color{black}
    \scriptsize
    \setlength{\tabcolsep}{8pt}
    \caption{Sensitivity to the archetype embedding fusion weight $\alpha$ on Mamo.C.}
    \label{tab:alpha_sensitivity}
    \begin{tabular}{c| c c c c c}
        \toprule
        $\alpha$
        & 0.30 & 0.50 & 0.55 & 0.70 & 0.90 \\
        \midrule
        \# Skills
        & 98 & 58 & 46 & 36 & 38 \\
        SA (\%)
        & 54.98 & 64.93 & 63.51 & 56.87 & 57.35 \\
        \bottomrule
    \end{tabular}
\end{table}

\begin{table*}[ht!]
    \centering
    \color{black}
    \scriptsize
    \setlength{\tabcolsep}{5pt}
    \caption{\textcolor{black}{
    Inference token consumption across five benchmarks for all
    compared methods. Token usage is reported in millions (M).
    }}
    \label{tab:cost_all_inference}
    \begin{tabular}{l|cccccc}
        \toprule
        Method
        & OptiBench
        & Mamo.C
        & OptMATH
        & IndustryOR
        & ComplexOR
        & Total \\
        \midrule
        Chain-of-Experts
        & 3.597M
        & 2.060M
        & 2.186M
        & 0.941M
        & 0.148M
        & 8.932M \\
        OptiMUS
        & 9.841M
        & 5.393M
        & 10.632M
        & 2.516M
        & 0.331M
        & 28.713M \\
        ORMind
        & 15.656M
        & 7.290M
        & 9.619M
        & 4.257M
        & 0.691M
        & 37.512M \\
        ORThought
        & 1.473M
        & 0.668M
        & 0.490M
        & 0.272M
        & 0.049M
        & 2.952M \\
        LEAN-LLM-OPT
        & 11.626M
        & 5.558M
        & 4.711M
        & 2.405M
        & 0.410M
        & 24.710M \\
        AlphaOPT
        & 6.524M
        & 3.233M
        & 2.980M
        & 1.474M
        & 0.271M
        & 14.482M \\
        Trace2Skill
        & 17.889M
        & 7.657M
        & 8.947M
        & 5.353M
        & 0.423M
        & 40.269M \\
        OptSkills
        & 27.524M
        & 11.585M
        & 12.820M
        & 7.030M
        & 0.676M
        & 59.635M \\
        \bottomrule
    \end{tabular}
\end{table*}

\begin{table*}[t]
    \centering
    \color{black}
    \scriptsize
    \setlength{\tabcolsep}{2.5pt}
    \caption{\textcolor{black}{
    Component-level token consumption of \textsc{OptSkills} during training.
    Percentages indicate the fraction of token consumption within
    each stage. Token usage is reported in millions (M).
    }}
    \label{tab:cost_training_breakdown}
    \begin{tabular}{lccccccccc}
        \toprule
        Stage
        & Samples
        & Extractor
        & \shortstack{Skill\\Selection}
        & \shortstack{Solver Pool\\Rollout}
        & \shortstack{Tool-calls\\Agent}
        & \shortstack{Skill\\Analysis}
        & \shortstack{Skill\\Build}
        & \shortstack{Skill\\Refine}
        & Total \\
        \midrule
        Clustering
        & 150
        & \shortstack{0.350M\\(3.9\%)}
        & \shortstack{0.000M\\(0.0\%)}
        & \shortstack{0.205M\\(2.3\%)}
        & \shortstack{7.488M\\(84.3\%)}
        & \shortstack{0.423M\\(4.8\%)}
        & \shortstack{0.411M\\(4.6\%)}
        & \shortstack{0.000M\\(0.0\%)}
        & 8.878M \\
        Learning
        & 150
        & \shortstack{0.372M\\(3.6\%)}
        & \shortstack{2.294M\\(22.1\%)}
        & \shortstack{0.016M\\(0.2\%)}
        & \shortstack{5.030M\\(48.4\%)}
        & \shortstack{1.556M\\(15.0\%)}
        & \shortstack{0.054M\\(0.5\%)}
        & \shortstack{1.063M\\(10.2\%)}
        & 10.384M \\
        \midrule
        Total
        & 300
        & \shortstack{0.722M\\(3.8\%)}
        & \shortstack{2.294M\\(11.9\%)}
        & \shortstack{0.221M\\(1.1\%)}
        & \shortstack{12.518M\\(65.0\%)}
        & \shortstack{1.979M\\(10.3\%)}
        & \shortstack{0.465M\\(2.4\%)}
        & \shortstack{1.063M\\(5.5\%)}
        & 19.262M \\
        \bottomrule
    \end{tabular}
\end{table*}

\begin{table*}[t]
    \centering
    \color{black}
    \scriptsize
    \setlength{\tabcolsep}{6pt}
    \caption{\textcolor{black}{
    Component-level token consumption of \textsc{OptSkills} during inference.
    Percentages indicate the fraction of token consumption within
    each benchmark. Token usage is reported in millions (M).
    }}
    \label{tab:cost_inference_breakdown}
    \begin{tabular}{lccccc}
        \toprule
        Dataset
        & Samples
        & Extractor
        & \shortstack{Skill\\Selection}
        & \shortstack{Tool-calls\\Agent}
        & Total \\
        \midrule
        OptiBench
        & 605
        & \shortstack{1.585M\\(5.8\%)}
        & \shortstack{8.677M\\(31.5\%)}
        & \shortstack{17.263M\\(62.7\%)}
        & 27.524M \\
        Mamo.C
        & 211
        & \shortstack{0.667M\\(5.8\%)}
        & \shortstack{3.077M\\(26.6\%)}
        & \shortstack{7.840M\\(67.7\%)}
        & 11.585M \\
        OptMATH
        & 166
        & \shortstack{0.635M\\(5.0\%)}
        & \shortstack{2.475M\\(19.3\%)}
        & \shortstack{9.710M\\(75.7\%)}
        & 12.820M \\
        IndustryOR
        & 100
        & \shortstack{0.284M\\(4.0\%)}
        & \shortstack{1.442M\\(20.5\%)}
        & \shortstack{5.305M\\(75.5\%)}
        & 7.030M \\
        ComplexOR
        & 18
        & \shortstack{0.053M\\(7.9\%)}
        & \shortstack{0.261M\\(38.6\%)}
        & \shortstack{0.362M\\(53.5\%)}
        & 0.676M \\
        \midrule
        Total
        & 1,100
        & \shortstack{3.224M\\(5.4\%)}
        & \shortstack{15.932M\\(26.7\%)}
        & \shortstack{40.480M\\(67.9\%)}
        & 59.635M \\
        \bottomrule
    \end{tabular}
\end{table*}

\textcolor{black}{
\textbf{Downstream Sensitivity to DBSCAN $\epsilon$.} We further evaluate the downstream sensitivity to the DBSCAN radius $\epsilon$ on Mamo.C. We vary $\epsilon$ from 0.01 to 0.15 and additionally include $\epsilon=0.20$ as a stress-test setting. As shown in Table~\ref{tab:dbscan_downstream_sensitivity}, performance remains relatively stable for $\epsilon \in [0.05, 0.15]$, with solving accuracy ranging from 63.51\% to 65.40\%. In contrast, performance drops at very small values of $\epsilon$ and at $\epsilon=0.20$. Small radii can over-fragment semantically related instances into separate clusters or noise points, weakening cross-instance aggregation during skill distillation, whereas an excessively large radius may merge distinct problem archetypes and produce less coherent skills. Overall, the intermediate range provides a reasonable trade-off between over-fragmentation and over-merging.
}

\textcolor{black}{
\textbf{Sensitivity to Archetype Embedding Fusion Weight $\alpha$.} We further evaluate the sensitivity to the fusion weight $\alpha$, which controls the relative contribution of ingredient and edited-problem embeddings in archetype construction. Table~\ref{tab:alpha_sensitivity} reports both the resulting number of skills and downstream solving accuracy on Mamo.C. Performance decreases at extreme values of $\alpha$. A smaller $\alpha$ places more weight on edited-problem embeddings, making clustering more sensitive to surface-level narratives and resulting in a more fragmented skill library, while a larger $\alpha$ relies more heavily on ingredient embeddings. Both $\alpha=0.50$ and $\alpha=0.55$ achieve strong performance. We use $\alpha=0.55$ because it provides comparable accuracy while producing a more compact skill library.
}

\subsection{\textcolor{black}{Computational Cost Analysis}}
\label{app:cost_analysis}

\textcolor{black}{
To complement the aggregate cost comparison in
Table~\ref{tab:cost_main}, we report the full benchmark-level
inference token consumption of all compared methods and further
decompose the cost of \textsc{OptSkills} by system component during
training and inference.
Table~\ref{tab:cost_all_inference} presents the inference token
consumption across all five benchmarks, while
Tables~\ref{tab:cost_training_breakdown} and
\ref{tab:cost_inference_breakdown} provide component-level
breakdowns for training and inference, respectively.
The component-level analysis separates the cost of extraction
from that of skill routing, trajectory generation, and
tool-based solving.
}

\textcolor{black}{
\textbf{Learning-stage breakdown.}
As shown in Table~\ref{tab:cost_training_breakdown}, the allocation
of computation differs substantially between the two training stages.
During clustering, the Tool-calls Agent accounts for 84.3\% of
token consumption, consistent with this stage primarily collecting
solving trajectories for subsequent skill distillation.
During Skill Learning, the computation becomes more distributed:
the Tool-calls Agent accounts for 48.4\%, while Skill Selection,
Skill Analysis, and Skill Refinement account for 22.1\%, 15.0\%,
and 10.2\%, respectively.
Across the full training process, the Tool-calls Agent remains
the largest contributor at 65.0\%, whereas the Extractor accounts
for only 3.8\%.
This shift reflects the different roles of the two stages:
initial skill discovery concentrates computation on collecting
solution evidence, while subsequent learning allocates a larger
fraction of the budget to selecting and updating reusable skills.
}

\textcolor{black}{
\textbf{Inference-stage breakdown.}
As shown in Table~\ref{tab:cost_inference_breakdown}, at test time,
the Tool-calls Agent accounts for 67.9\% of total token consumption,
followed by Skill Selection at 26.7\%, while the Extractor contributes
only 5.4\%.
Although the relative proportions vary across benchmarks, skill
routing and tool interaction consistently dominate the inference
budget.
The additional inference cost of \textsc{OptSkills} is therefore concentrated
mainly in selecting and applying skills through tool-based solving,
rather than in the extraction stage.
This identifies skill selection and tool interaction as the
primary targets for improving the efficiency of the current system.
}

\section{Case Study}
\label{appendix_casestudy}

To make the effect of skill injection more concrete, we include a representative case study in which the task is a cardinality-constrained pairwise selection problem with directed and asymmetric pairwise benefits. We compare a skill-guided trajectory with a no-skill trajectory on the same instance. Figure~\ref{fig:case_study} summarizes the two trajectories and highlights where the semantic error occurs.

\begin{center}
  \includegraphics[width=\columnwidth]{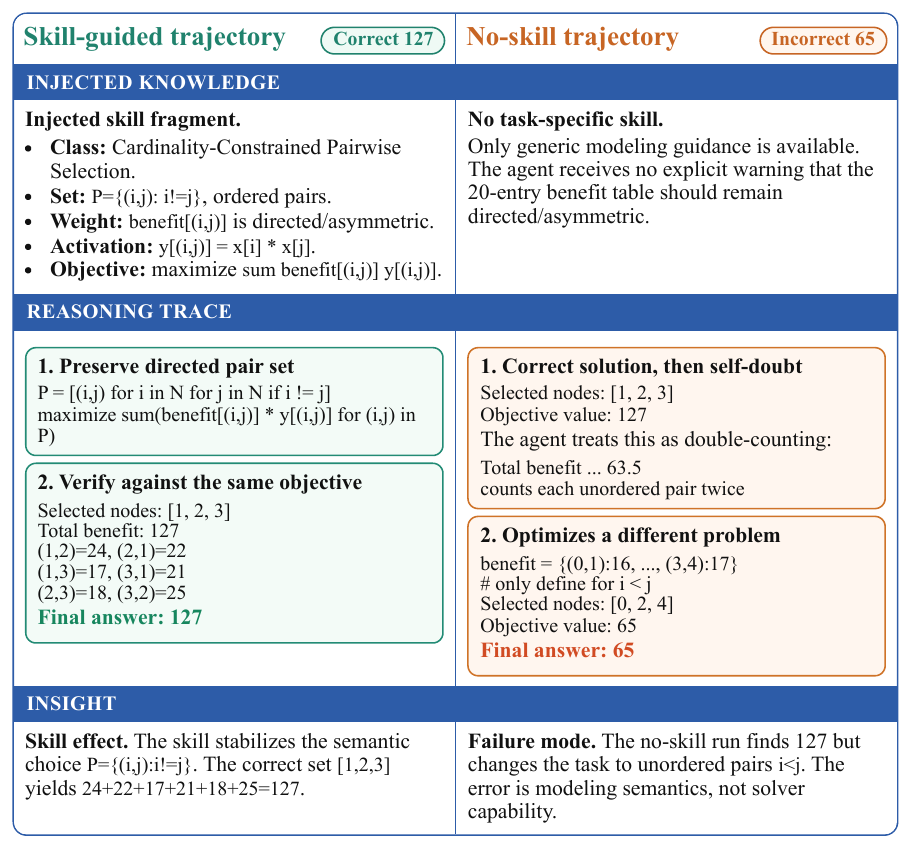}
  \captionof{figure}{
    \textbf{Case study of skill-guided optimization modeling.}
    A comparison of the skill-guided and no-skill trajectories on the same optimization instance.
    }
  \label{fig:case_study}
\end{center}

\section{Prompts}
\label{appendix_Prompts}
We provide the detailed prompts below.
\subsection{Prompt of LLM baseline}
\begin{prompttemplateBlue}{Prompt}
You are a practical function-calling optimization research agent. Follow the 3-stage guidance and rules below to complete the task:

### Modeling and solving the problem step by step
1. **Modeling Stage**: First output a valid formulation in JSON format wrapped in <formulation> tags.
2. **Solving Stage**: Generate the solver-based Python code and wrapped in <solver_code> tags.
    - you may use the python-based modeling language as follows: ortools, pyomo.
    - you may use the solver as follows: gurobi, clp, glpk, ipopt, highs, cbc, scip, mindtpy. You need to select an appropriate solver based on the type of the problem (linear or nonlinear, integer or continuous, and so on).
    - Solver code must print f"RESULT: {{objective_value}}".

### Problem Description
<problem_description>
{problem_description}
</problem_description>

### Output the result as follows:
<formulation>
{formulation}
</formulation>

<solver_code>
{solver_code}
</solver_code>
\end{prompttemplateBlue}

\subsection{Extractor Prompts}

\begin{prompttemplateBlue}{EXTRACTOR\_PROMPT\_EVAL}
You are an optimization-structure extractor.
Your goal is to ignore surface semantics and recover the underlying optimization structure.
Different stories may describe the same mathematical problem class. You must normalize them into the same structural representation whenever the variables, constraints, and objective are essentially the same.

You must do two things on the same problem:
1) extract scenario-agnostic optimization keywords into three slots;
2) minimally edit the problem text into a scenario-neutral version.

Task A: Keyword extraction
- Extract high-signal structural keywords and place them into:
  - `variable`
  - `constraint`
  - `objective`
- The goal is not only mathematical correctness, but also discriminative power: the extracted keywords should help distinguish different canonical optimization classes even when their algebraic forms are similar.
- The extracted keywords should capture the structural signature of the problem: the decision roles of variables, the coupling mechanisms in constraints, and the operational intent of the objective.
- Prefer canonical optimization roles, mechanisms, and structural motifs over generic algebraic descriptions.
- Prefer structurally discriminative keywords that best characterize the problem class, rather than the most generic mathematical-form label.
- Preserve distinctions between problem classes even when they share similar variable types, linear constraints, or objective form.
- If a keyword is needed, use lowercase snake_case and keep it scenario-agnostic.
- Use generic keywords such as variable type, linear equality/inequality, or lower/upper bound only when they add meaningful structural information or when no more specific structural keyword applies.
- Do not include a generic algebraic keyword when it is already implied by a more specific structural keyword, unless the generic keyword adds distinct information.
- For the `objective` slot, prefer keywords describing optimization intent, cost structure, or performance criterion over keywords describing only algebraic form.
- Do not output narrative entities unless they represent canonical optimization structure.
- Avoid redundant near-synonyms, overly broad labels, or multiple generic keywords when a more specific structural keyword better captures the optimization pattern.

Keyword list reference for inference:
<keywords_list>
{keywords_list}
</keywords_list>

When extracting keywords, prioritize canonical tokens from this list when they are semantically appropriate to the problem. Do not force a token from the list if none is a good fit.

Task B: Minimal scenario-neutral edit
- Produce `edited_problem` by preserving the original problem structure as much as possible.
- Keep mathematical meaning, quantities, equations/inequalities, and logical relations unchanged.
- Replace scenario-specific nouns/entities with generic placeholders where needed (for example: `entity_A`, `resource_i`, `task_j`).
- Do NOT simplify the optimization structure.
- Do NOT reorder, delete, or add constraints/objectives.
- Do not delete information.
- Do not summarize.
- Edit all scenario-specific background narratives, even if they carry no mathematical content.
- Keep this as a minimal edit not a paraphrase rewrite.

Confidence
- Return a confidence score in [0,1] for keyword extraction quality.
  - 0.8-1.0: Keywords are structurally accurate, scenario-agnostic, and clearly discriminative for the problem class;
  - 0.5-0.7: Most keywords are relevant, but some are generic or a few important structural distinctions are underspecified;
  - 0.2-0.4: Some keywords are relevant but require more inference, or important discriminative structure is missing;
  - 0.0-0.1: Few or no clear keywords can be extracted.

Problem description to analyze:
<problem_description>
{problem_description}
</problem_description>

Output JSON only:
{{
  "keywords": {{
    "variable": ["string"],
    "constraint": ["string"],
    "objective": ["string"]
  }},
  "edited_problem": "string",
  "confidence": 0.0
}}

Output constraints:
- Include all three keyword slots (`variable`, `constraint`, `objective`), each as an array (can be empty).
- `edited_problem` must be non-empty plain text.
- No extra keys.
- The confidence score is a float between 0.0 and 1.0.
- Return only JSON, no markdown/code fence/explanation.
\end{prompttemplateBlue}

\begin{prompttemplateBlue}{EXTRACTOR\_PROMPT}
You are an optimization-structure extractor.
Your goal is to ignore surface semantics and recover the underlying optimization structure.
Different stories may describe the same mathematical problem class. You must normalize them into the same structural representation whenever the variables, constraints, and objective are essentially the same.

You must do two things on the same problem:
1) extract scenario-agnostic optimization keywords into three slots;
2) minimally edit the problem text into a scenario-neutral version.

Task A: Keyword extraction
- Extract high-signal structural keywords and place them into:
  - `variable`
  - `constraint`
  - `objective`
- The goal is not only mathematical correctness, but also discriminative power: the extracted keywords should help distinguish different canonical optimization classes even when their algebraic forms are similar.
- The extracted keywords should capture the structural signature of the problem: the decision roles of variables, the coupling mechanisms in constraints, and the operational intent of the objective.
- Prefer canonical optimization roles, mechanisms, and structural motifs over generic algebraic descriptions.
- Prefer structurally discriminative keywords that best characterize the problem class, rather than the most generic mathematical-form label.
- Preserve distinctions between problem classes even when they share similar variable types, linear constraints, or objective form.
- If a keyword is needed, use lowercase snake_case and keep it scenario-agnostic.
- Use generic keywords such as variable type, linear equality/inequality, or lower/upper bound only when they add meaningful structural information or when no more specific structural keyword applies.
- Do not include a generic algebraic keyword when it is already implied by a more specific structural keyword, unless the generic keyword adds distinct information.
- For the `objective` slot, prefer keywords describing optimization intent, cost structure, or performance criterion over keywords describing only algebraic form.
- Do not output narrative entities unless they represent canonical optimization structure.
- Avoid redundant near-synonyms, overly broad labels, or multiple generic keywords when a more specific structural keyword better captures the optimization pattern.

Task B: Minimal scenario-neutral edit
- Produce `edited_problem` by preserving the original problem structure as much as possible.
- Keep mathematical meaning, quantities, equations/inequalities, and logical relations unchanged.
- Replace scenario-specific nouns/entities with generic placeholders where needed (for example: `entity_A`, `resource_i`, `task_j`).
- Do NOT simplify the optimization structure.
- Do NOT reorder, delete, or add constraints/objectives.
- Do not delete information.
- Do not summarize.
- Edit all scenario-specific background narratives, even if they carry no mathematical content.
- Keep this as a minimal edit not a paraphrase rewrite.

Confidence
- Return a confidence score in [0,1] for keyword extraction quality.
  - 0.8-1.0: Keywords are structurally accurate, scenario-agnostic, and clearly discriminative for the problem class;
  - 0.5-0.7: Most keywords are relevant, but some are generic or a few important structural distinctions are underspecified;
  - 0.2-0.4: Some keywords are relevant but require more inference, or important discriminative structure is missing;
  - 0.0-0.1: Few or no clear keywords can be extracted.

Problem description to analyze:
<problem_description>
{problem_description}
</problem_description>

Output JSON only:
{{
  "keywords": {{
    "variable": ["string"],
    "constraint": ["string"],
    "objective": ["string"]
  }},
  "edited_problem": "string",
  "confidence": 0.0
}}

Output constraints:
- Include all three keyword slots (`variable`, `constraint`, `objective`), each as an array (can be empty).
- `edited_problem` must be non-empty plain text.
- No extra keys.
- The confidence score is a float between 0.0 and 1.0.
- Return only JSON, no markdown/code fence/explanation.
\end{prompttemplateBlue}

\subsection{Executor Prompts}
\begin{prompttemplateBlue}{EXECUTOR\_SYSTEM\_PROMPT}
You are a practical function-calling optimization research agent. Follow the 3-stage guidance and rules below to complete the task:

### 3-Stage Guidance
1. **Modeling Stage**: First output a valid formulation in JSON format wrapped in <formulation> tags.
2. **Solving Stage**: Call the `run_code` tool to execute solver code. Ensure tool arguments are valid JSON.
3. **Final Stage**: Output the final numeric answer wrapped in <answer> tags.

### Core Rules
- At most one tool call per turn.
- Solver code must print `RESULT:<number>`.
- Never assume tool success without checking feedback.
- Select exactly one workflow from the skill (parallel alternatives, not sequential steps).
- Before any tool call, output: <formulation>{...valid JSON...}</formulation>
- After observing a valid numeric tool result, record the current best numeric candidate at the end of your assistant reply as <tmp>number</tmp>; <tmp> is temporary and does not terminate the task.
- If the tool output clearly reports the requested objective value in plain text but forgot to print RESULT:<number>, you may record that objective value as <tmp>number</tmp>; do not use <tmp> for infeasible, unknown, timeout, failed, or ambiguous outputs.
- If the tool output this turn gives the same numeric value, or reports an error, infeasible, unknown, timeout, or failed status, keep the previous <tmp>number</tmp> unchanged instead of clearing or replacing it.
- Final output must be numeric-only: <answer>number</answer>
- Only <answer>...</answer> can terminate the task; any intermediate JSON, <formulation>, or <tmp> is not final.

### Termination Guidance
- If the obtained result is the optimal value and conforms to the problem statement and the formulation, immediately output the final answer.
- If the obtained result is problematic or the `run_code` tool does not receive correct feedback, continue the process.
- If a feasible result is obtained but it is uncertain whether it is optimal, a verification tool call may be performed. If the verification is successful, immediately output the final answer; if there is a problem, continue the process.

### Task Context
<keywords>
{keywords}
</keywords>

<skill>
{skill}
</skill>
\end{prompttemplateBlue}

\begin{prompttemplateBlue}{EXECUTOR\_USER\_PROMPT}
### Task Description
<problem_description>
{problem_description}
</problem_description>
\end{prompttemplateBlue}

\subsection{Solver Selection Prompts}
\begin{prompttemplateBlue}{SOLVER\_SELECTION\_PROMPT}
You are selecting python solver library and backend combinations for optimization tasks.

Task:
- Choose top-k candidates from the provided solver catalog.
- Use problem description + structured keyword slots.
- Prefer backends that are likely to be available and stable.
- Catalog only contains framework/backend identifiers; detailed docs are loaded after selection.

Selection principles:
1. Infer problem family from objective/constraint keywords.
2. Prefer robust and standard backends when uncertain.
3. Keep diversity across frameworks/backends when possible.
4. Do not invent solver IDs; choose only from catalog.

Input:
<problem_description>
{problem_description}
</problem_description>

<keywords>
{keywords}
</keywords>

<top_k>
{top_k}
</top_k>

<solver_catalog_json>
{solver_catalog}
</solver_catalog_json>

Return JSON only:
{{
  "selected": [
    {{
      "solver_id": "string",
      "reason": "short reason"
    }}
  ]
}}    
\end{prompttemplateBlue}

\subsection{Skill Prompts}
\begin{prompttemplateBlue}{SKILL\_SELECTION\_PROMPT\_EVAL}
You are a skill selector used for skill routing. Route the most suitable skill for the current problem.

Your task:
- Select exactly one skill from the provided candidate list.
- You must choose only from the candidate `skill_id` values provided below.
- You must rely only on the candidate skill `name` and `description`.
- Do not use hidden assumptions about the full skill content.
- Do not invent a new skill.

Problem structure signals:
<keywords>
{keywords}
</keywords>

Scenario-neutral edited problem:
<edited_problem>
{edited_problem}
</edited_problem>

Candidate skills for the current active state:
<skill_candidates_json>
{skill_candidates_json}
</skill_candidates_json>

Return JSON only in this exact schema:
{{
  "skill_id": "string",
  "reason": "string",
  "confidence": 0.0
}}

Output constraints:
- `skill_id` must exactly match one candidate `skill_id`.
- `reason` must be brief and grounded in the provided keywords / edited problem / descriptions.
- `confidence` must be a float in [0.0, 1.0].
- No markdown, no code fence, no extra keys, no extra text.
\end{prompttemplateBlue}

\begin{prompttemplateBlue}{SKILL\_SELECTION\_PROMPT}
You are a conservative Operations Research skill selector used for online self-learning skill routing.

Your task:
- Decide whether an existing reusable skill is suitable for the current problem.
- If a suitable skill exists, select exactly one skill from the provided candidate list.
- If no candidate is structurally suitable, reject retrieval.
- You must choose only from the candidate `skill_id` values provided below when `decision` is `recall`.
- You must rely only on the candidate skill `name` and `description`.
- Do not use hidden assumptions about the full skill content.
- Do not invent a new skill.

Reject when:
- the candidate belongs to a different canonical optimization family,
- the current problem and candidate are not the same canonical optimization type,
- the core modeling structure, decision-variable pattern, objective, or constraint mechanism does not match,
- the match is based only on superficial keyword overlap,
- candidate descriptions are too generic to justify reuse,
- or you are uncertain that the selected skill is reusable for this problem.

Recall only when:
- the current problem and selected candidate strictly and clearly describe the same canonical optimization type,
- and the candidate description covers the current problem's core variable semantics, constraint mechanism, and objective semantics.

Strict decision policy:
- Trust your own structural judgment even when all candidates look imperfect.
- Do not force a recall just because a candidate is the closest available option.
- Do not recall a skill that would require adapting its canonical formulation to a different problem type.
- Reusable means directly applicable at the canonical type level, not merely adaptable.
- Be careful with problems that look similar at the surface level; similar keywords do not by themselves mean the same canonical optimization type.
- If the exact canonical type is not represented by any candidate description, choose `reject`.

Problem structure signals:
<keywords>
{keywords}
</keywords>

Scenario-neutral edited problem:
<edited_problem>
{edited_problem}
</edited_problem>

Candidate skills for the current active state:
<skill_candidates_json>
{skill_candidates_json}
</skill_candidates_json>

Return JSON only in this exact schema:
{{
  "decision": "recall",
  "skill_id": "string",
  "reason": "string",
  "confidence": 0.0
}}

Output constraints:
- `decision` must be exactly `recall` or `reject`.
- If `decision` is `recall`, `skill_id` must exactly match one candidate `skill_id`.
- If `decision` is `reject`, `skill_id` must be an empty string.
- `reason` must be brief and grounded in the provided keywords / edited problem / descriptions.
- `confidence` must be a float in [0.0, 1.0].
- No markdown, no code fence, no extra keys, no extra text.
\end{prompttemplateBlue}

\begin{prompttemplateBlue}{SKILL\_SELECTION\_PROMPT\_JUDGE}
You are a conservative Operations Research skill selector used for learning-stage skill routing.

Your task:
- Decide whether an existing reusable skill is suitable for the current problem.
- If a suitable skill exists, reuse exactly one skill from the provided candidate list.
- If no candidate is structurally suitable, route to new skill construction.
- You must choose only from the candidate `skill_id` values provided below when `decision` is `reuse`.
- You must rely only on the candidate skill `name` and `description`.
- Do not use hidden assumptions about the full skill content.
- Do not invent a new skill.

Reject when:
- the candidate belongs to a different canonical optimization family,
- the current problem and candidate are not the same canonical optimization type,
- the core modeling structure, decision-variable pattern, objective, or constraint mechanism does not match,
- the match is based only on superficial keyword overlap,
- candidate descriptions are too generic to justify reuse,
- or you are uncertain that the selected skill is reusable for this problem.

Reuse only when:
- the current problem and selected candidate strictly and clearly describe the same canonical optimization type,
- and the candidate description covers the current problem's core variable semantics, constraint mechanism, and objective semantics.

Strict decision policy:
- Trust your own structural judgment even when all candidates look imperfect.
- Do not force reuse just because a candidate is the closest available option.
- Do not reuse a skill that would require adapting its canonical formulation to a different problem type.
- Reusable means directly applicable at the canonical type level, not merely adaptable.
- Be careful with problems that look similar at the surface level; similar keywords do not by themselves mean the same canonical optimization type.
- If the exact canonical type is not represented by any candidate description, choose `new`.

Problem structure signals:
<keywords>
{keywords}
</keywords>

Scenario-neutral edited problem:
<edited_problem>
{edited_problem}
</edited_problem>

Candidate skills for the current active state:
<skill_candidates_json>
{skill_candidates_json}
</skill_candidates_json>

Return JSON only in this exact schema:
{{
  "decision": "reuse",
  "skill_id": "string",
  "reason": "string",
  "confidence": 0.0
}}

Output constraints:
- `decision` must be exactly `reuse` or `new`.
- If `decision` is `reuse`, `skill_id` must exactly match one candidate `skill_id`.
- If `decision` is `new`, `skill_id` must be an empty string.
- `reason` must be brief and grounded in the provided keywords / edited problem / descriptions.
- `confidence` must be a float in [0.0, 1.0].
- No markdown, no code fence, no extra keys, no extra text.
\end{prompttemplateBlue}

\begin{prompttemplateBlue}{SKILL\_ANALYSIS\_PROMPT}
You are an operations research trajectory analyst. Your goal is to generate reusable Standard Operating Procedures (SOPs) for OR skill construction. Your core task is to analyze exactly one rollout trajectory, and output reusable guidance in markdown blocks based on objective metrics.

Next, refer to the relevant keywords:
<keywords>
{keywords}
</keywords>

Next, refer to Objective Indicators
<Indicators>
{Indicators}
</Indicators>

Now, analyze the candidate data:
<trajectory>
{trajectory}
</trajectory>

Your output must be a single JSON object following this schema:
{
  "positive_sop": "### Modeling\\n- ...\\n\\n### Solving\\n- ...",
  "should_avoid": " "
}

Adhere to these output constraints:
- If `label` is "positive", `positive_sop` must be non-empty with sections "### Modeling" followed by "### Solving", containing reusable, scenario-agnostic bullet points. Set `should_avoid` to an empty string.
- If `label` is "negative", `should_avoid` must be non-empty, containing reusable, scenario-agnostic bullet points. Set `positive_sop` to an empty string.
- Do not include any text outside the JSON object.
- Keep It General: The skill should apply to similar problems, not just this one.
- Capture Executable Knowledge: When the trajectory includes effective code, extract the core logic as a reusable template. Good code templates are worth more than paragraphs of description.
- Brevity Matters: Aim for ~600 words. Focus on what's actionable.
\end{prompttemplateBlue}

\begin{prompttemplateBlue}{BUILD\_SKILL\_PROMPT}
You are an OR Skill Builder that writes workflow-oriented SKILL documents for long-term reuse.

Mission:
- Build one polished SKILL markdown from problem context and per-candidate analyses.
- The result must guide another model from formulation to executable solving code.
- Keep it practical, technically precise, and transferable.
- Follow the output template as a strict structural schema: do not add, remove, rename, reorder, or reformat any required section, heading level, or YAML frontmatter delimiter.

Design principles:
1. Distill lessons, do not narrate trajectory.
2. Keep workflow alternatives parallel (not sequential).
3. Use solver-aware modeling templates.
4. Make result parsing and failure handling explicit.
5. Prefer concise technical clarity over decorative prose.
6. Suitable code usage can be placed in every step with explicit comments.
7. Keep it general,Use placeholders instead of specific values. The skill should apply to similar problems, not just this one, not just this scenario. The Skill Name, Skill Description, Workflow Names, Step Names should all be generalized and not scenario-specific.

Hard format contract:
- Keep this section order exactly:
  frontmatter
  -> # Workflow 1 (...)
     -> ## Modeling stage
     -> ## Solving stage
  -> # Workflow 2 (...)
     -> ## Modeling stage
     -> ## Solving stage
- Do not add extra top-level sections.
- Do not output explanatory text outside the markdown.

Input:
<keywords>
{keywords}
</keywords>

<candidate_analyses_json>
{candidate_analyses}
</candidate_analyses_json>

Candidate analyses schema contract:
- Each item contains:
  - `candidate_id` (controller-provided identifier)
  - `label` (`positive` or `negative`)
  - `positive_sop` (markdown with `### Modeling` and `### Solving`)
  - `should_avoid` (markdown with `### Should Avoid`)
- Build skill content mainly from:
  - positive examples: `positive_sop`
  - negative examples: `should_avoid`
- Do not assume legacy fields such as `decision_basis`, `summary`, or nested step arrays.

Output markdown template (SKILL.md content only):
---
name: [skill_name]
description: |
  [One sentence covering both modeling and solving behavior.]
---

# Workflow 1 ([workflow_name])

## Modeling stage

### Strategy Overview
[Short paragraph.]

### Step 1 - [Model Step Name]
- [Action]
- [Action]

### Step 2 - [Model Step Name]
- [Action]
- [Action]

### Step ...

### Formulation Template
```json
{
  "sets": [],
  "parameters": [],
  "decision_variables": [],
  "objective": {
    "sense": "min",
    "expression": ""
  },
  "constraints": []
}
```

### Common Pitfalls
- [Pitfall]
- [Pitfall]

## Solving stage

### Strategy Overview
[Short paragraph.]

### Step 1 - [Solver Step Name]
- [Action]
- [Action]

### Step 2 - [Solver Step Name]
- [Action]
- [Action]

### Step ...

### Code Usage
```python
# build model from formulation
# solve with status / termination checks
```

### Common Pitfalls
- [Pitfall]
- [Pitfall]

# Workflow 2 ([workflow_name])

## Modeling stage

### Strategy Overview
[Short paragraph.]

### Step 1 - [Model Step Name]
- [Action]
- [Action]

### Step 2 - [Model Step Name]
- [Action]
- [Action]

### Step ...

### Formulation Template
```json
{
  "sets": [],
  "parameters": [],
  "decision_variables": [],
  "objective": {
    "sense": "min",
    "expression": ""
  },
  "constraints": []
}
```

### Common Pitfalls
- [Pitfall]
- [Pitfall]

## Solving stage

### Strategy Overview
[Short paragraph.]

### Step 1 - [Solver Step Name]
- [Action]
- [Action]

### Step 2 - [Solver Step Name]
- [Action]
- [Action]

### Step ...

### Code Usage
```python
# build model from formulation
# solve with status / termination checks
```

### Common Pitfalls
- [Pitfall]
- [Pitfall]

Quality bar:
- Workflows must be meaningfully different (backend/solver/API style).
- Avoid copy-paste duplication between Workflows.
- Keep placeholders generic and executable.
- Output starts with `---` and ends at markdown body.
\end{prompttemplateBlue}

\begin{prompttemplateBlue}{REFINE\_SKILL\_PROMPT}
You are a Skill Optimization Specialist. Your core task is to refine an existing operations research SKILL using one completed trajectory analysis and the current skill content, with the goals of improving the skill's effectiveness, preserving its reusable workflow quality, removing redundancy, generalizing specific cases, and enhancing structure.

Your goals are to:
- improve the SKILL's effectiveness,
- preserve and strengthen reusable workflow quality,
- remove redundancy,
- generalize only what is truly reusable,
- and maintain a stable, executable structure.

Core Execution Rules:
- Use the `<label>` as the supervision signal:
  - If `label` is `positive`: strengthen or clarify successful reusable procedures supported by the `<skill_analysis>`.
  - If `label` is `negative`: add or strengthen prerequisite checks, early warning signals, failure guards, and fallback guidance supported by the `<skill_analysis>`.
- Apply only the smallest set of edits justified by the trajectory analysis.
- Do not infer broad behavioral changes from a single narrow case.
- Preserve the SKILL's identity: do not over-generalize it into a broader template that weakens its intended optimization problem class, solver assumptions, modeling pattern, or retrieval specificity.

Input:
<skill>
{skill}
</skill>

<skill_analysis>
{skill_analysis}
</skill_analysis>

<label>
{label}
</label>

Refinement Constraints (Strictly Follow):
1. Preserve the outer document skeleton exactly:
   - keep the full YAML frontmatter,
   - keep both opening and closing `---`,
   - keep all existing YAML keys,
   - keep key order unchanged,
   - keep the top-level section structure and heading order unchanged unless explicitly impossible to maintain validity.

2. Within existing sections, you may rewrite, compress, deduplicate, clarify, and generalize content. Prefer local edits over full rewrites.

3. Preserve all useful existing workflows. Remove only weak, vague, repetitive, or redundant wording.

4. Do not remove explicit execution checks, solver/status verification steps, or parseable output contracts. You may clarify or tighten them if supported by the trajectory.

5. Exclude scenario-specific story details from the analysis, such as names, one-time events, or incidental numeric values, unless they represent a reusable operational threshold or rule.

6. Replace overly specific examples with reusable patterns when possible. Convert hardcoded task-specific values into placeholders such as `[TARGET]`, `[QUERY]`, `[TIME_LIMIT]`, `[CAPACITY]`, or similar appropriate abstractions.

7. Preserve skill-discriminative information. Do not remove or blur details that define this SKILL's distinctive optimization structure, including when relevant:
   - problem class,
   - decision variable pattern,
   - objective type,
   - key constraint structure,
   - solver or modeling assumptions.

8. Merge duplicate or near-duplicate content only when meaning is preserved. Consolidate overlapping explanations into clearer single statements.

9. For `positive` labels:
   - strengthen successful SOPs supported by the analysis,
   - clarify reusable step order,
   - make successful checks, thresholds, or decision criteria more explicit,
   - do not add unsupported new methods or techniques.

10. For `negative` labels:
   - prioritize prerequisite checks, early failure detection, recovery steps, and fallback guidance,
   - do not ban a method entirely unless the analysis clearly shows a general failure mode rather than a one-off execution issue,
   - do not overwrite or weaken proven successful workflows unless the analysis reveals a general flaw in them.

11. Do not introduce new solver strategies, modeling tricks, tools, or procedures unless they are clearly supported by either the existing SKILL or the trajectory.

12. Keep the content concise, actionable, and reusable. Remove verbosity that does not improve execution quality.

13. Maintain consistent formatting and scannability within the existing structure:
   - clear steps,
   - consistent bullet style,
   - concise wording,
   - no unnecessary prose expansion.

14. Perform a consistency audit across prose, formulation templates, and code examples:
   - ensure sign conventions, index ordering, objective sense, constraint direction, and variable semantics are mutually consistent,
   - if a contradiction exists, prefer the smallest edit that restores internal consistency,
   - do not preserve contradictory formulations merely to minimize edits.

Refinement Principles:
- Prefer minimal, evidence-based edits
- Generalize reusable patterns, not incidental details
- Preserve what makes the SKILL specifically useful
- Add safeguards carefully and only when justified
- Keep the SKILL executable, clear, and retrieval-distinctive

Output:
- Return only refined SKILL markdown, starting at the frontmatter (`---`).
- Do not output JSON.
- Do not output explanatory text.
- Do not wrap the answer in code fences.
\end{prompttemplateBlue}

\end{document}